\pdfoutput=1

\documentclass[11pt]{article}

\usepackage[final]{acl}

\usepackage{times}
\usepackage{latexsym}

\usepackage[T1]{fontenc}

\usepackage[utf8]{inputenc}

\usepackage{microtype}

\usepackage{inconsolata}

\usepackage{graphicx}
\usepackage{pdflscape}
\usepackage{makecell} 
\usepackage{multirow}
\usepackage{tikz}
\usetikzlibrary{trees, positioning}
\usepackage[edges]{forest}
\usepackage{tcolorbox}
\usepackage{booktabs}
\usepackage{amsmath}
\usepackage{adjustbox}
\usepackage[inline]{enumitem}
\usepackage{amssymb}  
\usepackage{cuted}      
\usepackage[skip=2pt]{caption}
\usepackage{subcaption}

\usepackage{stfloats}        
\usepackage[section]{placeins}
\usepackage{flafter}

\newcommand{\header}[1]{\vspace*{0.5mm}\noindent\textbf{#1}.}
\usepackage{array}
\newcolumntype{P}[1]{>{\raggedright\arraybackslash}p{#1}}

\title{A Comprehensive Taxonomy of Negation for NLP and Neural Retrievers}

\author{
Roxana Petcu, Samarth Bhargav, Maarten de Rijke, Evangelos Kanoulas \\
University of Amsterdam, The Netherlands \\
\texttt{\{r.m.petcu, s.bhargav, m.derijke, e.kanoulas\}@uva.nl}
}

\begin{document}
\maketitle

\begin{abstract}

Understanding and solving complex reasoning tasks is vital for addressing the information needs of a user. 
Although dense neural models learn contextualised embeddings,
they underperform on queries containing negation. To understand this phenomenon, we study negation in traditional neural information retrieval and LLM-based models. 
We (1) introduce a taxonomy of negation that derives from philosophical, linguistic, and logical definitions; (2) generate two benchmark datasets that can be used to evaluate the performance of neural information retrieval models 
and to fine-tune models for a more robust performance on negation; and (3) propose a logic-based classification mechanism that can be used to analyze the performance of retrieval models on existing datasets. Our taxonomy produces a balanced data distribution over negation types, providing a better training setup that leads to faster convergence on the NevIR dataset. Moreover, we propose a classification schema that reveals the coverage of negation types in existing datasets, offering insights into the factors that might affect the generalization of fine-tuned models on negation. Our code is publicly available on GitHub\footnote{\href{https://github.com/RoxanaPetcu/taxonomy-negation}{github.com/RoxanaPetcu/taxonomy-negation}}, and the datasets are available on HuggingFace\footnote{
\href{https://huggingface.co/datasets/RoxanaMaria/gpt4o-negation-controlled}{gpt4o-negation-controlled}} 
\footnote{\href{https://huggingface.co/datasets/RoxanaMaria/gpt4o-negation-free}{gpt4o-negation-free}}.

\end{abstract}
\section{Introduction}

A key factor contributing to accurate relevance in neural information retrieval (IR) systems, LLM-based re-rankers, and retrieval augmented generation (RAG) is acquiring language understanding capabilities through pre-training \cite{hosseini2021understanding}. Despite their extensive training setups, these models show persistent difficulty in handling negation \cite{mckenzie2024inversescalingbiggerisnt}, both in spoken and written language \cite{Ortega2016TheUS}. Negation is linguistically a complex phenomenon that, while guaranteed to be present in the training regime of any model, takes different forms depending on the task at hand. Human comprehension of negation comes as a result of understanding linguistic, morphological, and syntactic construction along with verbal cues (as defined in Appendix \ref{appendix:negation_properties}) and facial expressions \cite{Zuanazzi2023TrackingTB}. However, this multifaceted linguistic phenomenon is often reduced to a binary description in language processing systems: Does negation exist or not in a specific data set \cite{weller2024nevirnegationneuralinformation, zhang2024excluirexclusionaryneuralinformation}, and is it encoded or not by a model \cite{ravichander2022condaqa}. Addressing these discrepancies between human and system understanding of negation, we ask the following research questions:

\begin{enumerate}[label=(RQ\arabic*),nosep,leftmargin=*]
    \item Can we design a comprehensive taxonomy for negation?
    \item How can this taxonomy be applied to generate a more complete and balanced dataset?
    \item In what manner does model performance differ when fine-tuned on the taxonomy-driven dataset \\ versus prior existing datasets?
    \item How can this taxonomy be used to understand why models underperform on existing negation datasets?
\end{enumerate}

\begin{figure*}[!htbp]        
	\centering
	\includegraphics[width=0.9\linewidth]{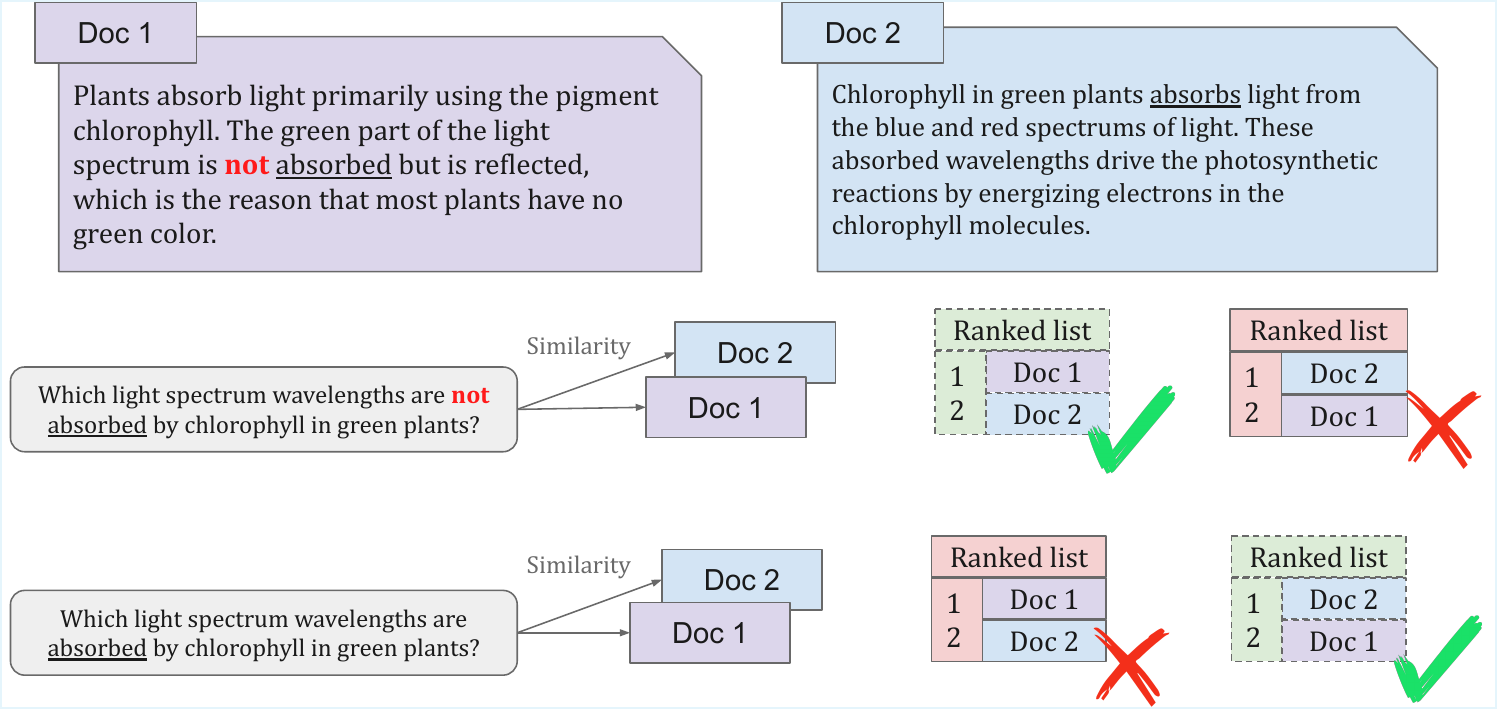}
	\caption{Example instance from our Free Generation dataset for sentential negation. Doc 1 is a passage retrieved from an existing Wikipedia article; Doc 2 is a minimally edited counterfactual whose truth value is flipped. The task is pairwise ranking. Given two queries that only differ in the presence of negation, the retrieval model must rank the corresponding document higher. The model succeeds if it ranks the correct document higher for both queries. There is a \(25\%\) random chance in pairwise accuracy.}
	\label{fig:example}
\end{figure*}

\noindent%
RQ1 aims to bring together research from the linguistic literature in a taxonomy on negation. We design our taxonomy to be exhaustive, with no overlap, and relevant to IR tasks. To address RQ2, we propose two synthetically generated datasets that cover all proposed negation types. Figure \ref{fig:example} illustrates the task alongside the data type represented in our datasets. RQ3 analyzes the performance of neural IR models, providing insight into the gap between human understanding and LLM encoding of negation. RQ4 connects the taxonomy to formalizations that can be used as data classification mechanisms, allowing to study existing datasets and identify reasons why fine-tuning does not guarantee a performance boost.
\section{Motivation}


\textbf{Negation has a long history in (computational) linguistics.} The study of opposition and its expression in the form of negation is a phenomenon that has been debated by, and provoked interest from linguists, logicians, metaphysicians, and philosophers \cite{seiver1944cicero, Horn1989-HORANH, kunen1987negation, halpern2005causes}. 
It is a highly complex expression of thought given its apparent simple form \cite{Horn1989-HORANH}. 
Other challenges are imposed by the ambiguity of the negation scope \cite{atlas1977negation}, and pragmatic inferences in conversational settings \cite{schloder2015pragmatic}.

\header{Proper treatment of negation is essential} Understanding negation is vital for retrieval models to provide the correct information to the user. Moreover, handling negation is vital to ensure that the retrieved generations are a correct response to the user query, since generated answers are particularly difficult to verify, as they cannot be grounded in established evidence \cite{wang2024factualitylargelanguagemodels}. Equally important is ensuring that RAG systems respect user-specified negation and avoid retrieving information the user explicitly does not search for.


\header{Fine-tuning on negation datasets} One could argue that this problem can be mitigated through fine-tuning \cite{Dolci2022FineTuningLM}. However, catastrophic forgetting occurs when a model is fine-tuned on a new dataset \cite{Hayes2019REMINDYN}, even if its distribution is similar to the original training data. In certain cases, fine-tuning can lead to a degradation of performance in the original training set \cite{peters2019tune, merchant2020happens}. Model sensitivity to parameter adjustments is particularly noticeable in information retrieval settings. This has been observed in traditional BERT-based architectures \citep{gerritse2022arjen} and LLMs \cite{soudani2024}. Although this behavior can be mitigated by freezing the model parameters and adding a language model head that is fine-tuned on a new dataset \cite{Huang2022FrozenCM, Lin2022FrozenCM}, this method restricts the capabilities the model can learn. \citet{weller2024nevirnegationneuralinformation} shows that fine-tuning on their proposed dataset (NevIR) leads to a noticeable decline in MSMarco generalization performance.

\header{Representations of negation} Another explanation for models under-performing on negation is 
under-representation of negation in crawled pre-training datasets \cite{hossain2020analysis}. 
An improper training can also be caused by the training objective. While contrastive loss pushes different content to be distant in the representation space, two negated statements are close in content while conveying opposite information. \cite{hosseini2021understanding, noji2020analysis} address the problem of misalignment between training objective and semantics by proposing an `unlikelihood' loss function to pre-train BERT on factually incorrect statements with negation cues. Recently, \cite{krasakis2025LSR} constructed compositional query representations to explicitly encode logical operators with Learned Sparse Retrieval (LSR), showing that penalizing negation in the query improves generalization.

\section{Related Work}

\header{Negation in IR}
Negation has been studied since early language models, e.g., \citet{jumelet2018negation} investigate the capabilities of LSTMs to locate the scope of negation, which they evaluate using a parse tree. Early work typically examines negation at the atomic sentence level. In contrast, negation in IR must be handled across pairs of queries and documents, as the presence of negation in a query can completely reverse the relevance of a document that otherwise is a semantic match. Therefore, IR systems must assess whether both the query and the document share the same polarity. i.e., positive or negative \cite{McQuire1998TheAO}. 
Negation in IR often takes the form of exclusion, which involves filtering information, and rejection of suggestions, which involves dismissing information, as mentioned by \citet{Tottie1991negation}. Having distinct types of negation poses an added challenge to defining it in an IR context, which can therefore be difficult and ambiguous.

\header{Negation in different modalities}
\citet{alhamoud2025visionlanguagemodelsunderstandnegation} propose a benchmark for understanding negation across 18 tasks and modalities spanning image, video, and medical data. Their experiments reveal that even with large-scale training, modern vision language models (VLMs) struggle with negation, often performing at random. The authors show that fine-tuning on large-scale synthetic datasets can approach a 10\% increase in performance. However, that forces the model to overfit on negation instead of making it reason on negation, as shown by achieving a good performance on one dataset but not generalizing on negation out of distribution \cite{zhang2020revisiting, zhou2021closer}.

\header{Retrieval models and LLMs for retrieval}
Information retrieval models evolved from lexical matching to dense retrieval, where the similarity between a query and documents is identified in a latent semantic space. These representations can be learned separately, i.e., with bi- and dual encoders, or together, i.e., with cross encoders. Dense models have been shown to outperform classical lexical matching in most scenarios \cite{karpukhin2020dense, khattab2020colbert}. In addition, LLMs are being fine-tuned to serve as the backbone of retrieval and ranking tasks \cite{zhu2023large}, bringing a boost in performance through their rich representations. LLM-based models used for retrieval are constructed on small-scale models, such as BERT \cite{devlin2019bert} and T5 \cite{raffel2020exploring}, or on larger-scale next token prediction models, such as Llama \cite{llama}, Mistral \cite{mistral} and Qwen \cite{qwen}. 

\header{Data generation using LLMs}
Data generation using LLMs has gained significant attention \cite{abolghasemi2024cause, askari2023expand, tunstall2023zephyr,abbasiantaeb2024let, liu2024chatqa}, and has been shown to be a viable method to expand the training dataset, improving performance in several tasks such as dialog generation~\cite{soudani2024surveyrecentadvancesconversational,askari2025selfseedingmultiintentselfinstructingllms}, reasoning \cite{yin2023alcunalargelanguagemodels}, negation \cite{li2023maqamultimodalqabenchmark} and exclusionary retrieval \cite{zhang2024excluirexclusionaryneuralinformation}.

\header{Existing negation datasets}
One of the first forays into negation understanding was in the medical domain, where research focused on automatically indexing clinical reports and discharge summaries \cite{Savova2010MayoCT, Niu2005AnalysisOP}. For example, Bio-Scope \cite{Zhu2019BioSCOPEFB} is a corpus of biomedical text mining that focuses on extracting accurate information on biological relations. Today, in the IR literature, we have access to publicly available datasets such as NevIR \cite{weller2024nevirnegationneuralinformation}, ExcluIR \cite{zhang2024excluirexclusionaryneuralinformation}, BoolQuestions \cite{Zhang2024BoolQuestionsDD}, Quest \cite{malaviya2023questretrievaldatasetentityseeking}, and RomQA \cite{zhong2022romqabenchmarkrobustmultievidence}. While these datasets contain logical operator annotations, the annotation system largely remains a single binary label for the presence of negation. 

\header{Research gap}
How is a taxonomy different from linguistic formalisations of negation in logic? Aristotle transferred the study of negation from the domain of ontology to logic and language \cite{smith2000aristotle}. The linguistic formalization of negation in logic defines how negation operates within formal systems \cite{Costa1974OnTT}, such as in classical logic, where a proposition \(p\) is negated through \(\neg p\) in which the truth value is flipped, or within modal and nonmonotonic logic \cite{Ketsman2020DatalogWN}, where it has more nuanced interpretations. In contrast, a taxonomy for negation would categorize different types and functions of negation in language and reasoning, such as lexical \cite{Staliunaite2020CompositionalAL} vs. semantical \cite{Urquhart1972SemanticsFR} negation, metalinguistic \cite{Horn1985MetalinguisticNA} vs. descriptive \cite{Miestamo2005StandardNT, Lee2017Chapter3M}, or negation as opposition \cite{Mettinger1994AspectsOS} vs. absence \cite{Faller2002SemanticsAP}. Although logic treats negation as a formal operation on truth values, a taxonomy explores its diverse roles in communication, cognition, and interpretation.

\section{Methodology}



We propose (1) a taxonomy for negation that is used to generate (2) two synthetic datasets that can be used for evaluating the performance of neural information retrieval models and for fine-tuning models to become more robust on negation, and (3) a classification mechanism that splits existing datasets into granular types of negation.

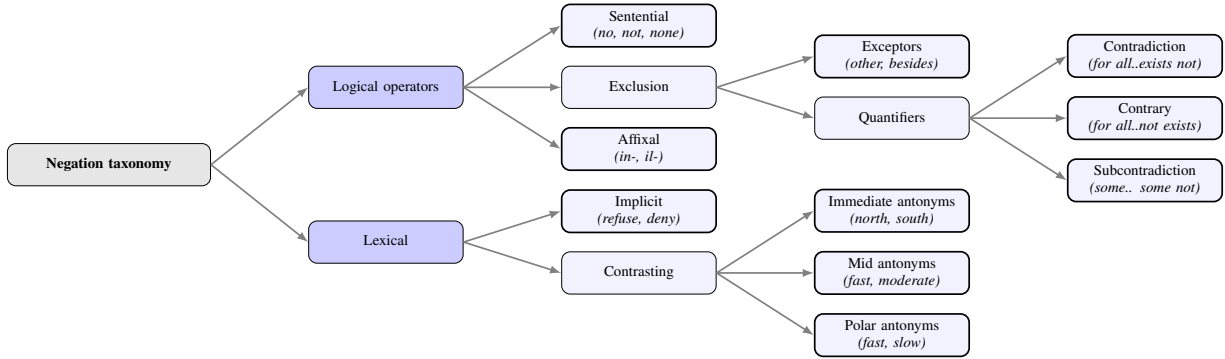
\begin{figure*}[ht]
	\centering
    \adjustbox{max width=1\textwidth}{%
	\begin{forest}
		for tree={
			grow'=0,
			rounded corners,
			draw,
			align=center,
			text centered,
			anchor=center,
			baseline,
			parent anchor=east,
			child anchor=west,
			edge={-latex, thick, draw=black!50},
			font=\scriptsize,
			fill=blue!5,
			minimum height=0.7cm,
			text width=2.4cm,
			l sep=16mm,
			s sep=3mm,
			inner sep=2pt,
		},
		[\textbf{Negation taxonomy}, fill=gray!20, text width=3.2cm
		[Logical operators, fill=blue!20
		[Sentential\\\textit{(no, not, none)}, draw=black, thick]
		[Exclusion
		[Exceptors\\\textit{(other, besides)}, draw=black,  thick] 
		[Quantifiers, tier=ExclusionBottom 
		[Contradiction\\\textit{(for all..exists not)}, draw=black,  thick]
		[Contrary\\\textit{(for all..not exists)}, draw=black,  thick]
		[Subcontradiction\\\textit{(some.. some not)}, draw=black,  thick]
		]
		]
		[Affixal\\\textit{(in-, il-)}, draw=black,  thick]
		]
		[Lexical, fill=blue!20
		[Implicit\\\textit{(refuse, deny)}, draw=black,  thick]
		[Contrasting
		[Immediate antonyms\\\textit{(north, south)}, draw=black,  thick]
		[Mid antonyms\\\textit{(fast, moderate)}, draw=black,  thick]
		[Polar antonyms\\\textit{(fast, slow)}, draw=black,  thick]
		]
		]
		]
	\end{forest}%
    }
	\caption{Negation taxonomy tree.}
	\label{fig:negation_taxonomy}
\end{figure*}

\subsection{Taxonomy}\label{section:taxonomy}

 

We derive our negation taxonomy from definitions in logic, philosophy \cite{Horn1989-HORANH} and natural language processing literature \cite{Tottie1991negation, McQuire1998TheAO}. Figure \ref{fig:negation_taxonomy} presents the taxonomy as a hierarchical tree, where each node denotes a negation type and its child nodes correspond to finer-grained subtypes. Table \ref{tab:negation_taxonomy} in Appendix \ref{appendix:taxonomy} includes query-document pairs exemplifying each negation type.

Our primary classification criterion is on the scope of negation (the part of a sentence whose meaning is altered by negation), distinguishing explicit negation realized by a logical operator \(\neg\) \cite{haegeman1995syntax}, from lexical negation that is present through the semantics of the word itself \cite{natayou2014explicit}. \textbf{Logical Operators} append to a word or clause, reversing its meaning. In \textbf{lexical} negation, a word or phrase inherently evokes negation, without the need for an appended operator.

We identify three types of logical operators based on literature review \cite{Horn1989-HORANH}. \textbf{Sentential} \cite{Zeijlstra2004SententialNA} negation is signalled by \underline{sentential operators} such as \textit{no}, \textit{not} and \textit{none}, which have a fixed syntactic role and occupy defined positions within a sentence. \textbf{Exclusion} \cite{MacCartney2008ModelingSC} is signalled by \underline{exclusionary operators} that are either \textbf{exceptors}, such as \textit{besides} and \textit{others} (exceptors represent a unique type of negation, see more in Appendix \ref{appendix:taxonomy}), or \textbf{quantifiers}, such as the universal quantifier \textit{for all} and the existential quantifier \textit{exists}. In Aristotelian logic \cite{Keenan1997GeneralizedQI, Horn1989-HORANH}, these quantifiers define three fundamental relations: \textbf{Contradiction}, \textbf{Contraries}, and \textbf{Subcontradiction}. Finally, \textbf{Affixal} \cite{Zimmer1966AffixalNI} negation is signalled by \underline{prefix} and \underline{suffix operators} that are preppended or appended to an existing word, such as: \textit{un-}, \textit{in-}, \textit{im-}, \textit{il-}, \textit{ir-}, \textit{dis-}, \textit{non-}, \textit{mis-}, \textit{ill-}, \textit{-less}, \textit{-free} \cite{Wahyuni2014ANAO}.

We identify two types of lexical negation. \textbf{Implicit} \cite{Madva2016WhyIA} negation is composed of words that are inherently negative through their meaning, e.g.: \textit{refuse}, \textit{deny}, \textit{exclude}, \textit{reject}, \textit{avoid}, \textit{lack}, \textit{fail}. \textbf{Contrasting} \cite{Trillas2017AntonymsNA} negation is composed of words that convey negation in pairs, but are not negative independently. These can be called contrasting pairs of antonyms. \textbf{Immediate} antonyms are opposite words with no degree of variation between them; \textbf{Polar} antonyms are opposite words with degrees of variation between them, and \textbf{Mid} antonyms represent samples from the interpolation of two polar antonyms. For more special cases of negation that we do not cover in this study, see Appendix \ref{appenfix:cool_negation_types}.


\subsection{Data Generation}
We generate two synthetic datasets designed to cover all negation types described in the taxonomy. We construct the datasets as follows: (1)~we prompt an LLM to generate 100 \textit{topics} of general knowledge to ensure familiarity \cite{askari2025selfseedingmultiintentselfinstructingllms} and avoid long-tail knowledge; (2)~for each topic, we ask the LLM to return one \textit{Wikipedia page} that we check using the Wikipedia API, ensuring the generations are grounded in documented and factual information; 
(3)~conditioned on a Wikipedia page, the LLM generates pairs \((q_1, doc_1)\) and \((q_2, doc_2)\) following the template of CondaQA \cite{ravichander2022condaqa} and NevIR  \cite{weller2024nevirnegationneuralinformation}. (3.1)~Given detailed prompts constructed for the individual negation type, we ask the LLM to retrieve a paragraph that contains one specific negation as defined in the taxonomy. If the document does not contain explicit markers for the specified negation, the model will retrieve the closest match and rephrase it by injecting specific markers, i.e., keywords such as \textit{impossible} instead of \textit{not possible}. This phenomenon was observed with affixal negations, which our approach translated as a sentential one, as they are guaranteed to be semantically equivalent. The other types of negation that were not always present in the documents were the quantifiers, which can be translated from one to the other with logic transformations. (3.2)~Given the extracted paragraph, the LLM generates a query. 
This is the process of generating one pair \((q_1, doc_1)\). (3.3)~For generating the second pair, we employ two strategies to produce different degrees of lexical overlap between the negated datasets. (1) \textbf{Free Generation}: generate a positive query \(q_2\) by removing the negation from \(q_1\); generate a positive document \(doc_2\) by answering \(q_2\). (2) \textbf{Controlled Generation}: generate a positive query \(q_2\) by removing the negation from \(q_1\); generate a positive document \(doc_2\) by removing the negation from \(doc_1\). The two synthetically generated datasets have \(1505\) and \(1479\) instances, respectively, where a single instance has pairs \((q_1, doc_1)\) and \((q_2, doc_2)\). Appendix \ref{appendix:prompts_for_generation} provides the prompts used for generation, and an additional verification step for guaranteeing the relevance of documents; Table \ref{tab:model_statistics} and Figure \ref{fig:class_distr} summarize the dataset statistics and distribution of generated labels.

	
	

\subsection{LM Logic classification}\label{sec:logic_classification}

Negation can be analysed at two granularities. \textbf{Sentence-level:} some negation types can be identified at the sentence level; if two sentences are either both negative or both positive, the pair agrees in polarity \cite{mahany2022negation}, and if they do not, it conveys a negative polarity relationship (sentential, exclusionary, affixal, and implicit). \textbf{Pair-level:} the negation polarity can only be identified by comparison, i.e., whether both statements can be true at the same time (quantifiers and contrasting negation). We propose a classification mechanism that assigns each instance in an existing dataset a category outlined in our taxonomy by converting it to natural logic using typed lambda  \((\lambda)\) calculus formalisations \cite{Barendregt1985TheLC} (see Appendix \ref{appendix:taxonomy}). We generate formalisations for each instance by prompting a model with an instruction to generate the typed lambda calculus proof, and return the predicates, quantifiers and \(\lambda\)-typed formula. 
We categorize an existing dataset in four iterative steps:

\begin{description}[leftmargin=0cm]

	\item[\textbf{Step 1: Predicate Classification}]  
	We check the returned predicates. If any predicate defined in the deconstruction of the query is of sentential, exclusionary, affixal, or implicit nature (as classified by the LLM), we label the instance accordingly. Since they are sentence-level negations, we only study the queries.
	
	\item[\textbf{Step 2: Quantifier Pattern Matching}]  
	If no predicates are found, we analyse query and document pairs. We extract the logical quantifiers present in both the query and document (both pairs, see Appendix \ref{appendix:lm_logic_classification}), and check if any of the following logical patterns are identified as contradiction, contrary and subcontradition definitions \cite{Horn1989-HORANH}: \((\forall \dots \exists \neg)\), \((\forall \dots \neg \exists)\), \((\exists \dots \exists \neg)\). Instances matching any of these patterns are labelled accordingly.
	
	\item[\textbf{Step 3: Semantic Antonyms Detection}] 
    We will assume the only other potential negation is both at the semantic level and only detectable in paired interactions (in contrast, a predicate such as \textit{refuse} inherently carries a negative polarity, whereas a predicate such as \textit{slow} does not). We check such antonym pairs with the \textit{nltk} library.
	
	\item[\textbf{Step 4: Absence of Negation}]  
	If none of the previous conditions are met, we conclude that the instance does not contain negation according to our taxonomy.
	
\end{description}




\section{Experimental Setup}
Throughout this study, we use the GPT-4o-mini model \cite{openai2024gpt4ocard} to conduct experiments that aim to answer our research questions. More precisely, we evaluate retrieval models to reveal the necessity of our taxonomy-driven synthetic data, evaluate categorized existing datasets to show the usefulness of our logic-driven mechanism, and fine-tune to show that a coverage of negation types can help with generalisation.

\begin{table*}[!t]
	\centering
        {\small
	\setlength{\tabcolsep}{4pt} 
	\renewcommand{\arraystretch}{1.2} 
	
	\begin{tabular}{llllrl}
		\toprule
		\textbf{Model} & \textbf{Architecture} & \textbf{Training objective} & \textbf{Training dataset} & \textbf{Size} & \textbf{Tokenizer}  \\
		\midrule
		BM25 & Sparse & Retrieval & N/A & N/A & N/A \\
		DPR [\citenum{karpukhin2020dense}] & Bi-Encoder & Retrieval & NQ & 219M & BERT \\
		coCondenser [\citenum{cocodensor}]& Bi-Encoder & Retrieval & MSMarco & 110M & BERT \\
		Dragon [\citenum{dragon}]& Bi-Encoder & Retrieval & MS MARCO & N/A & BERT  \\
		msmarco-bert-base-dot-v5 & Dual Encoder & Semantic Search & MSMarco & 110M & BERT \\
		multi-qa-mpnet-base-dot-v1 & Dual Encoder & Semantic Search& QA & 110M & MPNet \\
		Sentence-T5 & Dual Encoder & Sentence Similarity & NLI & 220M & T5 \\
		ColBERTv1 [\citenum{khattab2020colbert}]& Late Interaction & Retrieval & MSMarco & 110M & BERT \\
		ColBERTv2 [\citenum{santhanam2021colbertv2}] & Late Interaction & Retrieval & MSMarco & 110M & BERT  \\
		MonoT5 Base [\citenum{nogueira-etal-2020-document}]& Crossencoder & Ranking & MSMarco & 223M & T5 \\
		MonoT5 Large [\citenum{nogueira-etal-2020-document}]& Crossencoder & Ranking & MSMarco & 737M & T5 \\
		MonoT5 3B [\citenum{nogueira-etal-2020-document}]& Crossencoder & Ranking & MSMarco & 2.85B & T5 \\
		stsb-roberta-large & Crossencoder & Sentence Similarity & STS-B & 355M & RoBERTa \\
		qnli-electra-base & Crossencoder & NLI & QNLI & 110M & ELECTRA \\
		nli-deberta-v3-base & Crossencoder & NLI & MultiNLI, SNLI & 184M & DeBERTa \\
		Qwen2-1.5B-Instruct [\citenum{qwen}]& Transformer & NTP & Crawled  & 1.5B & Qwen2Tokenizer \\
		Qwen2-7B-Instruct [\citenum{qwen}]& Transformer & NTP & Crawled & 7B & Qwen2Tokenizer \\
		Mistral-7B-Instruct [\citenum{mistral}]& Transformer & NTP & Crawled & 7B & BPE \\
		Llama-3.1-3B-Instruct [\citenum{llama}]& Transformer & NTP & Crawled  & 7B & Llama \\
		Llama-3.2-8B-Instruct [\citenum{llama}]& Transformer & NTP & Crawled  & 7B & Llama \\
		\bottomrule
	\end{tabular}
    }
	\caption{Model comparison for our experiments. NLI refers to natural language inference, and NTP refers to next token prediction. byte pair encoding with fallback. The crawled datasets represent undefined large training sets.}
	\label{tab:model_comparison}
\end{table*}

\header{Evaluation of the generation} We assess the quality of the generated datasets with human annotation on 5\% of the generations, with two annotators evaluating each instance on: 
(1) relevance of documents to each query, (2) presence of negation, (3) naturalness, (4) coherence, and (5) consistency of information within the document. The annotation was conducted with LabelStudio.\footnote{https://labelstud.io/} We assess the annotations on quantitative and qualitative measures, together with the annotator agreement. Appendix \ref{appendix:annotators} illustrates the questions for the annotators, metrics used, alongside further details for the setup. For both performance and inner annotator agreement, we use metrics such as f1-score, average on ordinal scales, and (weighted) Cohen's Kappa. Tables \ref{tab:performance_annotators} and \ref{tab:inner_agreement} report the annotation metrics. The main findings are as follows:

\begin{itemize}[leftmargin=*,nosep]
    \item Annotators reported 71--77\(\%\) accuracy for document relevance and 83\%--88\% f1 score for negation presence.
    \item On a scale of 1--5, the annotators reported an approximate quality of 4 on naturalness, coherence, and consistency of language.
    \item The inner annotator agreement passed significance values for sentential and contrasting negation. For implicit and quantifiers, the test shows borderline agreement in language quality.
    \item The biggest disagreement was noticed in the exceptors.
    \item Human performance on the synthetic datasets shows a pairwise accuracy score of \(0.6571 \pm 0.0202\) for free generation, and \((0.6643 \pm 0.0101\) for controlled generation on identifying the relevant document for each question.
\end{itemize}

\header{Evaluation of the classification mechanism} We evaluate the quality of our classification mechanism by assessing it against the generated datasets, for which we have access to golden labels by design of construction: we generate data for each type of negation conditioned on a taxonomy-dependent prompt. We run the classification mechanism on the free generation dataset, and obtain a balanced accuracy score of \(86.84\%\) and an F1 score of \(86.95\%\). We notice that around \(54\%\) of missclassifications are contrary negations missclassified as contradictions. In our experiments, all models perform similarly between these two types of negation, as they are logically and lexically very similar. Therefore, we assume it does not affect our study.

\header{Retrieval Models} We study the performance of lexical, bi-encoder, cross-encoder, late interaction and transformer models trained for first-stage retrieval, ranking, sentence similarity, natural language inference (NLI) and next token prediction (NTP). We follow the experimental setup introduced by \citet{elsen2025reproducing}. We show the specifications of all models in Table \ref{tab:model_comparison}. 

\header{Datasets} We evaluate on three benchmarks. \textbf{NevIR} and \textbf{ExcluIR} are two contrastive benchmarks where each instance comprises of two documents and two queries that only differ by a targeted negation, or exclusion. We also use \textbf{MSMarco} \textit{dev} partition, which is not specifically designed for contrastive pairs, but is used simply as a complex retrieval benchmark.

\header{Metrics} The metric used to evaluate the task is \textbf{pairwise accuracy}: for each instance queries \(q_1, q_2\) and documents \(d_1, d_2\), the model independently ranks \(\{d_1, d_2\}\). The prediction is correct only when the system places \(d_1\) above \(d_2\) for \(q_1\) and inverts the order for \(q_2\). Random performance for pairwise accuracy is \(25\%\).

\header{Fine-tuning} We fine-tune three models: ColBERTv1, multi-qa-mpnet-base-dot-v1, and Mistral-7B-Instruct for 20 epochs on the free generated dataset and evaluate on NevIR \cite{weller2024nevirnegationneuralinformation} test and MSMarco \cite{nguyen2016ms} dev data.

\begin{figure}[h] 
	\centering
	\includegraphics[width=\linewidth]{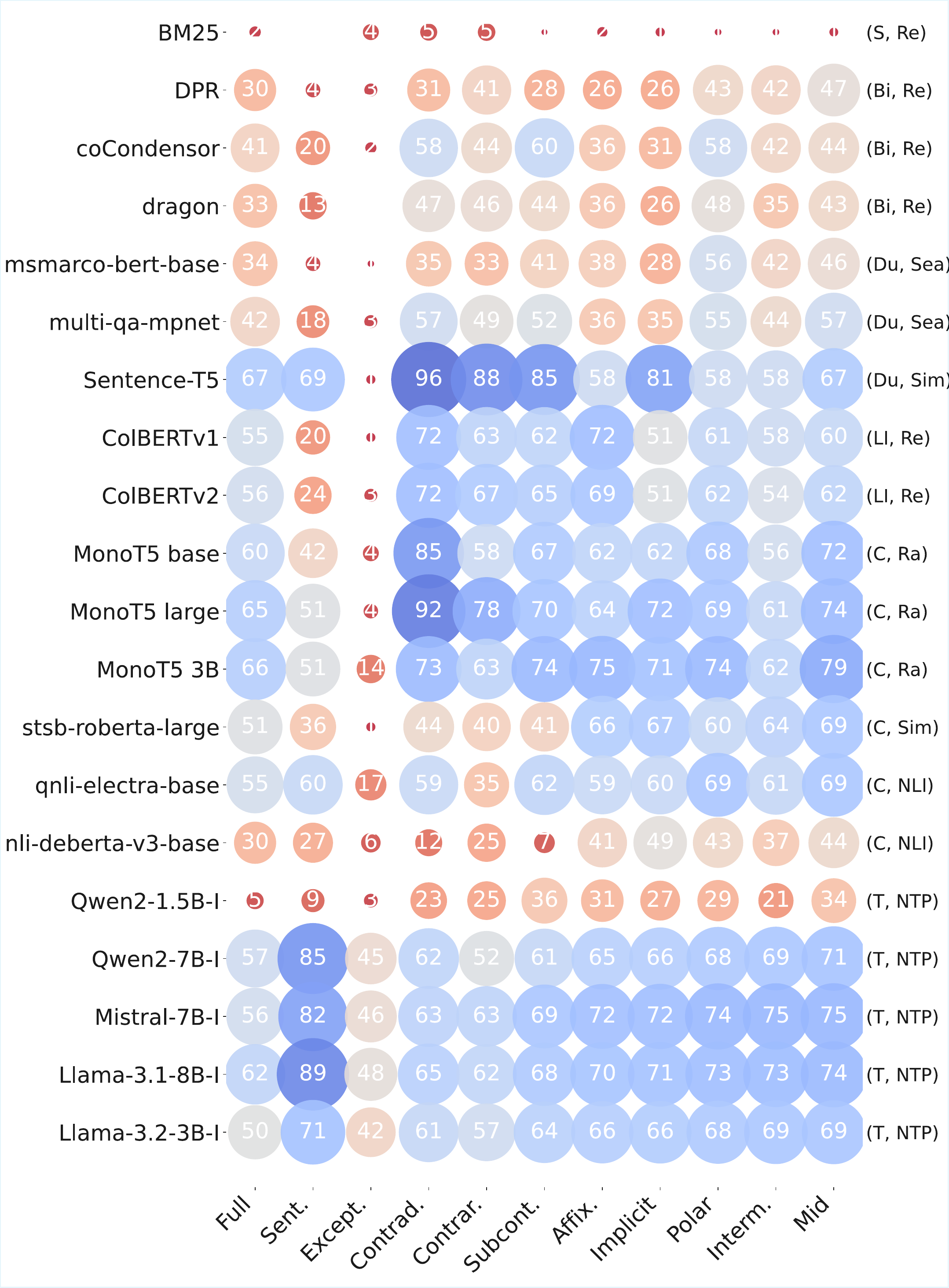}
	\caption{Pairwise Accuracy on the free generations dataset. The first result column contains the full dataset; later columns represent one negation type each. Models are represented by the rows, where \textbf{I} is a shortcut for \textbf{Instruct}. On the right, we assign labels expressing the architecture and training objective of each model: the first position shows the architecture, i.e., \textbf{S}parse, \textbf{Bi}-encoder, \textbf{Du}al encoder, \textbf{C}rossencoder, and \textbf{T}ransformer; the second position shows the training objective, i.e., \textbf{Re}trieval, \textbf{Sea}rch, \textbf{Sim}ilarity, \textbf{Ra}nking, \textbf{N}atural \textbf{L}anguage \textbf{I}nference, and \textbf{N}ext \textbf{T}oken \textbf{P}rediction. For a close-up, see Appendix \ref{ap:results}.}
	\label{fig:free_gen_pairwise_acc}
\end{figure}

\section{Results}

Our experiments are designed to investigate the following hypotheses: (H1) some negation types are better encoded in the model internal representations than others, (H2) model specifics such as architecture, training objective, size and backbone significantly influence performance on negation, (H3) existing datasets have an uneven representation on negation, (H4) fine-tuning on our synthetically generated dataset will show systematic improvement in the downstream task presented in Figure \ref{fig:example}.

\subsection{Evaluation on Synthetic Data}



Figure \ref{fig:free_gen_pairwise_acc} illustrates 20 models evaluated on the free generation dataset. Sparse, dual, and biencoders exhibit poor performance on all types of negation, except Sentence-T5: a dual encoder trained for semantic similarity. Both late-interaction and all cross-encoder models, except nli-deberta-v3-base, show strong performance on all negation types. BERT and T5-based cross-encoders perform better than models with a RoBERTa, ELECTRA, and DeBERTa backbone. All transformer-based models, except for Qwen 1.5B (which has a disadvantage in size, and which has been trained for NTP) perform well on almost all negation types. 

We perform a one-way ANOVA to test the significance of the results. ON model architecture, the ANOVA test reports a p-value of \(1.0087e-11\), and the Tukey HSD shows a significant difference between sparse and dense models. When grouping on the training objective, ANOVA indicates \(p = 1.5709e-04\), with significant differences between combinations of NTP, retrieval, and semantic search, and between sentence similarity vs. retrieval. The test shows a statistically significant difference between exceptors and all other types of negation. The experiments confirm hypothesis H1 and H2, that is, some negation types are better encoded than others, and that model specifics, such as architecture and training objective, influence performance. An analysis on the controlled generation dataset is illustrated in Figure \ref{fig:controlled_gen_pairwise_acc} in Appendix \ref{ap:results}, where a similar behavior is seen; however, the patterns are even stronger, with a general trend toward higher performance. This can be inherent in the data generation process, i.e., document 2 is generated by changing the negation in document 1 (as compared to directly answering query 2).



\begin{figure}[!t]
	\centering
	\includegraphics[width=0.8\linewidth]{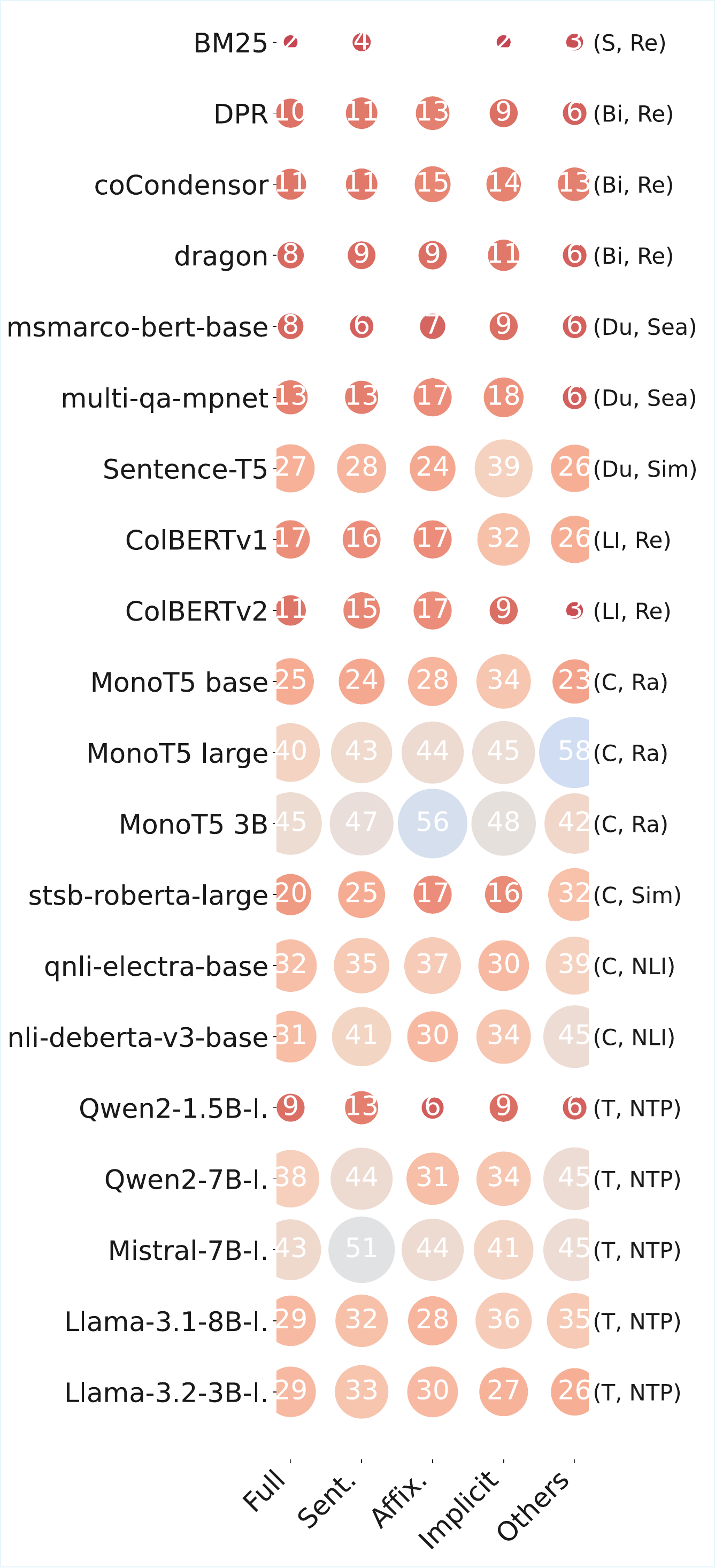}
	\caption{Pairwise Accuracy on NevIR as split with our classification mechanism.}
	\label{fig:nevirpa}
\end{figure}

\subsection{Evaluation on Logic Filtered NevIR}

When we apply the classification mechanism on the validation set of NevIR, we find that three main types of negation are present. Out of 225 pairs, \(\{79,\allowbreak\ 54,\allowbreak\ 44\}\) correspond to \(\{\text{Sentential},\allowbreak\ \text{Affixal},\allowbreak\ \text{Implicit}\}\), while \(31\) have been classified as not containing negation, in which case we label as \(\text{Others}\), while the remaining \(17\) pairs are spread across the other types of negation present in the taxonomy. These results are in line with hypothesis H3, which states that existing datasets have an uneven distribution of negation types.

Figure \ref{fig:nevirpa} shows that models perform worse on the NevIR dataset compared to our synthetically generated dataset. Sentence-T5 exhibits the best performance among bi- and dual-encoders. ColBERTv1 has a higher performance than ColBERTv2, and the MonoT5 models perform the best on all types of negation. Similarly to Figure \ref{fig:free_gen_pairwise_acc}, we notice that the performance in all models for sentential negation is higher than affixal or implicit. Qwen2-1.5B performs the worst of all LLMs, similarly to synthetic experiments.

\subsection{Evaluation on Logic Filtered ExcluIR}

When applying the classification mechanism on the ExcluIR test set, we find three types of negation: \(\{\text{Sentential},\allowbreak\ \text{Exclusionary},\allowbreak\ \text{Implicit}\}\) with \(\{189,\allowbreak\ 2820,\allowbreak\ 113\}\) pairs out of \(3452\). Moreover, \(297\) have been classified as ``Other'' while \(32\) are distributed among the other classes. This means that more than \(81\%\) of the entire dataset has been classified as exclusionary. These results further support hypothesis H3.

As shown in Figure \ref{fig:excluirpa} (Appendix \ref{ap:results}), the performance of the model is approximately uniform between the three identified types of negation. This finding contradicts with our synthetic data experiments, where exclusionary negation was significantly more difficult to encode than the other types of negation. To further inspect the source of this discrepancy, we take a closer inspection of the ExcluIR instances identified as ``Sentential'' or ``Implicit''. This reveals that these instances only have a different rephrasing of a task that essentially is still exclusion. 
One example extracted from the dataset is \textit{`Can you tell me about Paul Ziert's involvement in founding the Bart Conner Gymnastics Academy in Norman, Oklahoma, while avoiding any mention of Bart Conner's role in the academy?'}. Our categorization mechanism identifies this instance as ``Implicit'', while it has the form of a set subtraction, as per the definition of exceptors.

\subsection{Fine-tuning}

We fine-tune ColBERTv1, multiqa-mpnet-base-dot-v1, and Mistral-7B-Instruct on the free generation dataset, NevIR, and a mixed strategy with both datasets. We evaluate the finetuned models against NevIR dev set and MSMarco dev small. 

\noindent\textbf{Train partitions:} The NevIR training set is composed of 1,896 triplets. The train partition of our synthetically generated dataset consists of 2,114 triplets. When fine-tuning mixed data, we have a total of 2,005 triplets.

\noindent\textbf{Evaluation partitions:} We evaluate against the test partition of NevIR that has 2.8k triplets (2 triplets = 1 pair), and against the dev partition of MSMarco.

\begin{table}[t]
  \centering
  {\small
  \setlength{\tabcolsep}{5pt}      
  \renewcommand{\arraystretch}{1.2} 
  \begin{tabular}{ll*{3}{c} *{3}{c}}
    \toprule
     &  & \multicolumn{3}{c}{\textbf{NevIR} P.Acc. $\uparrow$}  & 
     \multicolumn{3}{c}{\textbf{MSMarco} MRR@10$\uparrow$}\\
    \cmidrule(lr){3-5} \cmidrule(lr){6-8}
      & & E1 & E6 & E20 
        & E1 & E6 & E20 \\
    \midrule
    \multirow{3}{*}{\rotatebox[origin=c]{90}{ColBERT}}
      & NevIR     & .21 & .24 & \underline{.45}
                 & .37 & \textbf{.37} & \textbf{.34}        \\
      & Synth    & \underline{.23} & \underline{.33} & .36       
                 & .36 & .34 & .31         \\
      & Mixed    & \textbf{.23} & \textbf{.40} & \textbf{.48}     
                 & \textbf{.37} & .33 & .31 \\
    \midrule
    \multirow{3}{*}{\rotatebox[origin=c]{90}{MultiQA}}
      & NevIR     & .12 & \underline{.51} & \textbf{\underline{.52}}
                 & \textbf{.35} & \textbf{.17} & \textbf{.06}     \\
      & Synth     & \underline{.34} & .38 & .40       
                  & .33 & .07 & .03         \\
      & Mixed     & \textbf{.36} & \textbf{.52} & .50    
                  & .26 & .03 & .01   \\
    \midrule
    \multirow{3}{*}{\rotatebox[origin=c]{90}{Mistral}}
      & NevIR     & \underline{.70} & \underline{.78} & \underline{.78}
                & .53 & .58 & \textbf{.60}  \\
      & Synth     & .58 & .58 & .58         
                 & \textbf{.59} & .55 & .55     \\
      & Mixed     & \textbf{.72} & \textbf{.78} & \textbf{.78}     
                 & .57 & \textbf{.60} & .54 \\
    \bottomrule
  \end{tabular}
  }
  \caption{Results for ColBERT, MultiQA and Mistral when trained on NevIR, Synth and Mixed data, and evaluated on NevIR and MSMarco. Columns E0, E1, E6, E20 represent epochs 0 (before backprop.), 1, 6 and 20; P. Acc. stands for pairwise accuracy, while MRR@10 for mean reciprocal rank at 10.}
  \label{tab:finetuning}
\end{table}


\subsubsection{Evaluation on NevIR}
As shown in Table \ref{tab:finetuning} and in Figure \ref{fig:finetuning_nevir} in Appendix \ref{appendix:finetuning}, fine-tuning \textbf{ColBERT} and \textbf{MultiQA} on our synthetic dataset yields an immediate performance gain on the NevIR development set, however peaking while fine-tuning on NevIR train reaches higher performance in the last epoch. This is to be expected as for the synthetic data we evaluate OOD. To assess in-distribution performance, we apply mixed fine-tuning by combining the two datasets and shuffling the data. The model achieves high performance significantly faster than when simply fine-tuned on NevIR, giving the overall best performance. \textbf{Mistral} shows the same behaviour with mixed fine-tuning. This supports hypothesis H4, that our synthetically generated dataset helps in capturing negation. Overall, we notice that fine-tuning on our synthetic data brings a quick performance boost against the NevIR dev and test sets, indicating that our proposed datasets capture the notion of negation.


\subsubsection{Evaluation on MSMarco}
When evaluated against MSMarco (Table \ref{tab:finetuning} and Figure \ref{fig:colbert_finetuning_eval_msmarco} in Appendix \ref{appendix:finetuning}), we notice that the generalizability of \textbf{ColBERT} and \textbf{MultiQA} drops when fine-tuned on any dataset. Interestingly, \textbf{Mistral} displays a more stable fine-tuning process; however, adding synthetic data drops performance even further. Although MSMarco generalization is known to be negatively affected when models are fine-tuned out of distribution, our results show a trade-off: synthetic and mixed training helps generalisation in the negation domain, but it further harms generalisation on MSMarco.

\section{Conclusion}

In this study, we propose a philosophy, logic and linguistic-grounded taxonomy for negation along two synthetic datasets that can be used for evaluating existing neural retrieval, ranking and LLM reranker models, and for fine-tuning models to increase their capabilities on negation. Through our study, we found that (1) cross-encoders and LLM rerankers are better at encoding negation, (2) NevIR and ExcluIR have a limited coverage of negation types, and (3) fine-tuning on our synthetic datasets helps performance in a negation domain.

These insights confirm that negation is a complex phenomenon and that a thorough taxonomy brings advantages as a starting point for generating fine-tuning data. The taxonomy-based classification of current datasets, together with model evaluation, shows that having a broad coverage of negation types is vital. Our fine-tuning experiments confirm that the synthetic datasets bring a performance boost; however, it also indicates that fine-tuning data might not be the sole factor behind model difficulty with negation. The training objective and architectural backbone play a big role in model performance performance. However, different training objectives are a promising direction for future work. Moreover, we propose investigating negation in a retrieval setting with a large corpora. Moreover, while generalization drops with fine-tuning, we propose investigating the training objective by applying reinforcement learning on negation with a small subset, similar to R1-Search \cite{jin2025searchr1trainingllmsreason}. 

\clearpage
\section*{Limitations}
Our work proposes a new dataset for investigating negation and improving performance in a negation setting, and a filtering mechanism for studying existing datasets. However, there are certain limitations to our study. Our dataset is limited to a binary classification redefined as a pairwise ranking task, and therefore is not directly applicable to a ranking setting with a large corpus. Moreover, the data is generated using GPT-4o mini. While the faithfulness of information is not the direct scope of this paper, having a more controlled generation process would be beneficial. Lastly, a broader study on datasets such as BoolQuestions, RomQA and Quest would offer a more extensive study.

\section*{Acknowledgments}
The evaluation of our generated data was done through LabelStudio. Moreover, we acknowledge our colleagues who helped with human evaluation and annotation: Panagiotis Eustratiadis, Jasmin Kareem, Clara Rus, David Vos, Maria Heuss, Lu Zhang, and Catherine Chen. We also want to acknowledge Maria Aloni, who offered help and feedback for our linguistic study. 

This research was (partially) supported by the Dutch Research Council (NWO), under project numbers 024.004.022, NWA.1389.20.\-183, and KICH3.LTP.20.006, the European Union under grant agreements No. 101070212 (FINDHR) and No. 101201510 (UNITE), and
Ahold Delhaize.
Views and opinions expressed are those of the author(s) only and do not necessarily reflect those of their respective employers, funders and/or granting authorities.

\bibliography{references}

\clearpage
\appendix
\section{Appendix}

This appendix offers further material that supports the study. It is organised as follows: Appendix \ref{appendix:negation_properties} defines the properties of negation that are briefly referenced in the study. Appendix \ref{appendix:taxonomy} gives an example in an information retrieval style for each type of negation present in the taxonomy, alongside further definitions of exceptors and typed lambda calculus. Appendix \ref{appendix:prompts_for_generation} lists all the prompts used to generate the datasets. Appendix \ref{appenfix:cool_negation_types} mentions use cases that we do not explicitly account for in this study, although they are interesting to study. \ref{appendix:lm_logic_classification} lists details into applying the categorization mechanism on the ExcluIR dataset. Appendix \ref{appendix:annotators} includes the survey that the human annotators completed to perform a qualitative evaluation of the generated data. Appendix \ref{ap:results} contains the results of evaluating the models against the controlled generated dataset and the ExcluIR data. Finally, Appendix \ref{appendix:statistical_analysis} offers a statistical analysis of the annotator's answers.

\subsection{Negation Properties}\label{appendix:negation_properties}

Drawing inspiration from \citet{ConanDoyle2012negation}, we define the following properties of negation:
 
\begin{itemize}

    \item \textbf{Negation cues}: Negation cues can be single words, multiwords, prefixes, such as im-, or suffixes, such as -less. They introduce the negation in the sentence.

    Example: \textit{She did \textbf{not} go to the movies, but went to the theater instead.}
    
    \item \textbf{Negated event}: The main event or property that is being negated. For example, if we define $\neg$ as a negation operation, i.e. $\neg A$, then A is the negated event. 
    
    Example: \textit{She did \textbf{not} \underline{go} to the movies, but went to the theater instead.}
    
    \item \textbf{Negated scope}: Extension of the negated event; part of the sentence where the negation propagates and changes its semantics. The parts of the sentence that are not affected by negation should be left out the scope. 

    Example: \textit{She did \textbf{not} \underline{go to the movies}, but went to the theater instead.}
    
\end{itemize}

\subsection{Taxonomy}\label{appendix:taxonomy}

In this section, we give a definition of exceptors using set operations, supporting our claim that exceptors are inherently a different type of negation compared to the rest of the taxonomy. This difference might influence how models perform on this negation type. We also give a definition of typed lambda calculus. Moreover, we provide examples for each negation type present in the taxonomy in the movie domain to exemplify the negation types in a retrieval setting. The examples are illustrated in Table \ref{tab:negation_taxonomy}.

\begin{table*}[ht]
	\centering
	\resizebox{\linewidth}{!}{
		\begin{tabular}{lllllll}
			\toprule
			\rotatebox{90}{\hspace*{-3.5mm}\textbf{Scope}} & 
			\begin{tabular}{@{}l}\textbf{Negation} \\ \textbf{category} 
            \end{tabular} & 
			\begin{tabular}{@{}l}\textbf{Negation} \\ \textbf{subcategory} 
            \end{tabular} & 
			\begin{tabular}{@{}l}\textbf{Aristotelian} \\ \textbf{logic}
            \end{tabular} & 
			\textbf{Examples} & 
			\textbf{Level} \\ 
            \midrule
			\multirow{13}{*}{\rotatebox{90}{Logical operators}} &
			\begin{tabular}{@{}l@{}}Sentential \\ \textit{(no, not, none)} \end{tabular} & & &
			\begin{tabular}{@{}l@{}} 
            Q: Movies that \textbf{do not} feature Tom Hanks. \\ D: Forrest Gump  features Tom Hanks. 
            \end{tabular} & Sentence \\
            \cmidrule{2-6}
			& \multirow{8}{*}{\begin{tabular}{@{}l@{}}Exclusion \end{tabular}} &
			\begin{tabular}{@{}l@{}} Exceptors \\ \textit{(others, besides} \\ \textit{but, except)} \end{tabular} & &
			\begin{tabular}{@{}l@{}}
            Q: Movies with Tom Hanks \textbf{besides} Forrest Gump. \\ 
            D: Forrest Gump is a widely acclaimed movie. \end{tabular} & Sentence \\ 
			\cmidrule{3-6}
			&  & \multirow{6}{*}{\begin{tabular}{@{}c@{}} Quantifiers \end{tabular}} &
			Contradiction & \begin{tabular}{@{}l@{}}
            Q: What are \textbf{all} movies with Tom Hanks? \\ 
            D: Here are \textbf{some} movies \textbf{without} Tom Hanks.. \end{tabular} & Pair
			\\ 
            \cmidrule{4-6}
			&  &  &
			Contrary & \begin{tabular}{@{}l@{}} 
            Q: What are \textbf{all} movies with Tom Hanks? \\ 
            D: There \textbf{exist no} movies with Tom Hanks. \end{tabular} & Pair
			\\ 
            \cmidrule{4-6}
			&  &  &
			Subcontradiction & \begin{tabular}{@{}l@{}} 
            Q: What are \textbf{some} movies with Tom Hanks? \\ 
            D: Here are \textbf{some} movies \textbf{without} Tom Hanks. \end{tabular} & Pair
			\\ 
            \cmidrule{2-6}
			& Affixal & & &
			\begin{tabular}{@{}l@{}} 
            Q: What are some movies with \textbf{un}happy endings? \\ D: These movies have happy endings. \end{tabular} & Sentence \\ 
            \midrule
			\multirow{7}{*}{\rotatebox{90}{Lexical}} &
			Implicit & & &
			\begin{tabular}{@{}l@{}} Q: Are there any movies with Tom Hanks \\ that \textbf{failed} people's expectations?. \\ D: This movie succeeded in public's eye. \end{tabular} & Sentence \\  
            \cmidrule{2-6}
			& \multirow{6}{*}{\begin{tabular}{@{}l@{}} Contrasting \end{tabular}} &
			Immediate Antonyms & & \begin{tabular}{@{}l@{}} 
            Q: A movie that is \textbf{professional}. \\ D: This is a \textbf{casual} movie. \end{tabular} & Pair
			\\ 
            \cmidrule{3-6}
			&  & Mid Antonyms & & \begin{tabular}{@{}l@{}} 
            Q: Movie where Tom Hanks is running very \textbf{fast}. \\ D: In this movie, Tom Hanks runs \textbf{moderately paced}. \end{tabular} & Pair
			\\ 
            \cmidrule{3-6}
			&  & Polar Antonyms & & \begin{tabular}{@{}l@{}} 
            Q: Movie where Tom Hanks is running very \textbf{fast}. \\ 
            D: In this movie, Tom Hanks runs very \textbf{slow}. \end{tabular} & Pair
			\\ 
            \bottomrule
		\end{tabular}
	}
	\caption{The proposed taxonomy of negation categories and their formalization.}
	\label{tab:negation_taxonomy}
\end{table*}

\textbf{Exceptors} represent a unique type of negation. While the other negation types take the form of opposition, i.e., two propositions \(p\) and \(\neg p\) cannot be true at the same time, exceptions are a form of set subtraction. More precisely, if we denote a domain S = \{\text{all candidate answers}\}, an exception set \(E \subseteq S = \{\text{items to exclude}\}\) and an exclusionary query \(Q_{\text{ex}} = S \setminus E\), then any document D that satisfies the exclusionary query \(Q_{\text{ex}}\) will inherently satisfy the whole set \(S\) as a consequence of \(S \setminus E\ \subseteq S\).

\textbf{Typed lambda calculus} is a formal system that decomposes any statement into a logic form, by defining abstract predicates and determiners, either assuming their truth value, or reaching unit clauses that can only be True or only False (reaching a contradiction). The primary goal of typed lambda calculus is to provide a framework for meaning composition with flexible functions (predicates and determiners).

\subsection{Data Generation}\label{appendix:prompts_for_generation}

\begin{figure*}[!ht]
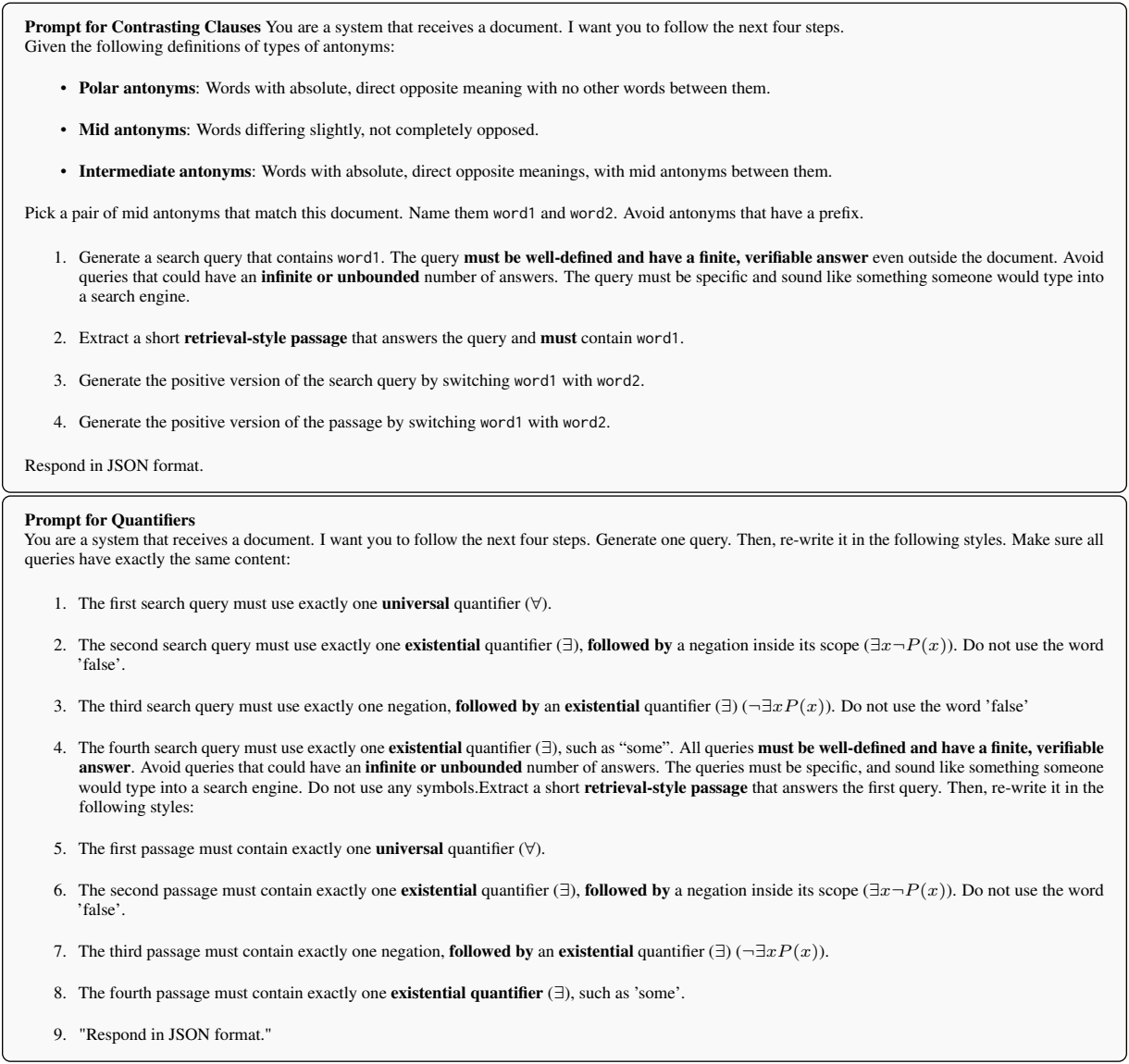

  \centering
  \scriptsize
    \begin{subfigure}{\textwidth}
	\begin{tcolorbox}[colback=gray!5!white, colframe=black!75!black, boxrule=0.5pt, arc=3pt, left=6pt, right=6pt, top=4pt, bottom=4pt]
		
		\textbf{Prompt for Contrasting Clauses}
        You are a system that receives a document. I want you to follow the next four steps.
        
        Given the following definitions of types of antonyms:
        \begin{itemize}
            \item \textbf{Polar antonyms}: Words with absolute, direct opposite meaning with no other words between them.
            \item \textbf{Mid antonyms}: Words differing slightly, not completely opposed.
            \item \textbf{Intermediate antonyms}: Words with absolute, direct opposite meanings, with mid antonyms between them.
        \end{itemize}
		
            Pick a pair of mid antonyms that match this document. Name them \texttt{word1} and \texttt{word2}. Avoid antonyms that have a prefix.
            
            \begin{enumerate}
                \item Generate a search query that contains \texttt{word1}. The query \textbf{must be well-defined and have a finite, verifiable answer} even outside the document. Avoid queries that could have an \textbf{infinite or unbounded} number of answers. The query must be specific and sound like something someone would type into a search engine.
                
                \item Extract a short \textbf{retrieval-style passage} that answers the query and \textbf{must} contain \texttt{word1}.
                
                \item Generate the positive version of the search query by switching \texttt{word1} with \texttt{word2}.
                
                \item Generate the positive version of the passage by switching \texttt{word1} with \texttt{word2}.
                
            \end{enumerate}
            
            Respond in JSON format.

	\end{tcolorbox}
    \end{subfigure}
    \begin{subfigure}{\textwidth}
        	\begin{tcolorbox}[colback=gray!5!white, colframe=black!75!black, boxrule=0.5pt, arc=3pt, left=6pt, right=6pt, top=4pt, bottom=4pt]
		
		\textbf{Prompt for Quantifiers}
		
		You are a system that receives a document. I want you to follow the next four steps. Generate one query. Then, re-write it in the following styles. Make sure all queries have exactly the same content:
		
		\begin{enumerate}
			\item The first search query must use exactly one \textbf{universal} quantifier (\(\forall\)).
			\item The second search query must use exactly one \textbf{existential} quantifier (\(\exists\)), \textbf{followed by} a negation inside its scope (\(\exists x \neg P(x)\)). Do not use the word 'false'.
			\item The third search query must use exactly one negation, \textbf{followed by} an \textbf{existential} quantifier (\(\exists\)) (\(\neg \exists x P(x)\)). Do not use the word 'false'
			\item The fourth search query must use exactly one \textbf{existential} quantifier (\(\exists\)), such as ``some''. All queries \textbf{must be well-defined and have a finite, verifiable answer}. Avoid queries that could have an \textbf{infinite or unbounded} number of answers. The queries must be specific, and sound like something someone would type into a search engine. Do not use any symbols.Extract a short \textbf{retrieval-style passage} that answers the first query. Then, re-write it in the following styles:
			\item The first passage must contain exactly one \textbf{universal} quantifier (\(\forall\)).
			\item The second passage must contain exactly one \textbf{existential} quantifier (\(\exists\)), \textbf{followed by} a negation inside its scope (\(\exists x \neg P(x)\)). Do not use the word 'false'.
			\item The third passage must contain exactly one negation, \textbf{followed by} an \textbf{existential} quantifier (\(\exists\)) (\(\neg \exists x P(x)\)).
			\item The fourth passage must contain exactly one \textbf{existential quantifier} (\(\exists\)), such as 'some'.
			\item "Respond in JSON format."
			
		\end{enumerate}
	\end{tcolorbox}
    \end{subfigure}
  \caption{Prompts for Contrasting Clauses and Quantifiers}
  \label{fig:contrasting_prompts}
\end{figure*}

In this section, we show the prompts used for generating the synthetic datasets for free and controlled generation. We illustrate the prompt for generating sentential negation in Figure \ref{fig:sentential-negation-prompt}. The prompts for generating exceptors, affixal and implicit negation are similar, where only steps 1 and 2 are different. We illustrate steps 1 and 2 for each of these negation types in Figure \ref{tab:prompt-variants}. The prompts for contrasting clauses and quantifiers are shown in Figure \ref{fig:contrasting_prompts}.

\begin{figure*}[!ht]
  \centering
    \begin{subfigure}{\linewidth}
	\begin{tcolorbox}[colback=gray!5!white, colframe=black!75!black, boxrule=0.5pt, arc=3pt, left=6pt, right=6pt, top=4pt, bottom=4pt]
	
	\textbf{Prompt for Sentential Negation}

	 You are a system that receives a document. I want you to follow the next four steps:
	
	\begin{enumerate}
		\item Generate a search query that contains exactly \textbf{one} negation word ('no', 'not', or 'none').\\
			It should \textbf{not} be accompanied by a quantifier. \\
			The query \textbf{must be well-defined and have a finite, verifiable answer} even outside the document. Avoid queries that could have an \textbf{infinite, unbounded or exhaustive} number of answers.\\
			Also, avoid queries that have the answer 'yes' or 'no'.\\
			The query must be specific, and sound like something someone would type into a search engine.
		\item Extract a short \textbf{retrieval-style passage} that contains exactly \textbf{one} negation word ('no', 'not', or 'none').\\
	   		- If the passage \textbf{does not contain} a negation, add exactly \textbf{one} negation word ('no', 'not', or 'none').
		\item Generate the positive version of the search query by removing the negation.
		\item Generate the positive version of the passage by removing the negation. Keep the other words intact. 
            \item Respond in JSON format.
		\end{enumerate}
    \end{tcolorbox}
    \end{subfigure}
  \caption{Prompts for Sentential Negation}
  \label{fig:sentential-negation-prompt}
\end{figure*}

\begin{figure*}[!ht]
	\centering
	\begin{minipage}{\linewidth}
		\centering
            {\small 
		\renewcommand{\arraystretch}{1.3} 
		\setlength{\tabcolsep}{8pt} 
		\begin{tabular}{p{1.5cm} p{13.3cm}}
			\hline
			\textbf{Variant} & \textbf{Differences in Step 1 and Step 2} \\
			\hline
			\textbf{Sentential} & Step 1: Generate a query that contains exactly \textbf{one} negation word ('no', 'not', or 'none'). It should \textbf{not} be accompanied by a quantifier. 
			The query \textbf{must be well-defined and have a finite, verifiable answer} even outside the document. Avoid queries that could have an \textbf{infinite, unbounded or exhaustive} number of answers.
			Also, avoid queries that have the answer 'yes' or 'no'.
			The query must be specific, and sound like something someone would type into a search engine. 
			\newline Step 2: Extract a short \textbf{retrieval-style passage} that contains exactly \textbf{one} negation word ('no', 'not', or 'none').\
			- If the passage \textbf{does not contain} a negation, add exactly \textbf{one} negation word ('no', 'not', or 'none'). \\
			\hline
			\textbf{Exceptor} & Step 1:  Generate a search query that contains exactly \textbf{one} exclusionary word such as ('others', 'besides', 'but', or 'except'). The query \textbf{must be well-defined and have a finite, verifiable answer} even outside the document. Avoid queries that could have an \textbf{infinite or unbounded} number of answers. The query must be specific, and sound like something someone would type into a search engine.
			\newline Step 2: Extract a short \textbf{retrieval-style passage} that answers the query. Make sure the passage does \textbf{not} contain an exclusionary word such as ('others', 'besides', 'but', or 'except'). Make sure the passage also contains the excluded part from the query. \\
			\hline
			\textbf{Affixal} & Step 1: Generate a search query that contains exactly \textbf{one} affixal negation such as ('un-', 'in-', 'im-', 'il-', 'ir-', 'dis-', 'non-', 'mis-', 'ill-'). An affixal negation adds a prefix or suffix to reverse the meaning of a word. The query should not contain any other negation. The query \textbf{must be well-defined and have a finite, verifiable answer} even outside the document. Avoid queries that could have an \textbf{infinite or unbounded} number of answers. The query must be specific, and sound like something someone would type into a search engine. \newline Step 2: Extract a short \textbf{retrieval-style passage} that answers the query. - In answering the query, the passage must contain exactly \textbf{the same} affixal negation as in the query. - If the passage \textbf{does not contain} an affixal word, add exactly \textbf{the same} one as in the query. The passage should not contain any other negation.\\
			\hline
			\textbf{Implicit} & Step 1: Generate a search query that contains exactly \textbf{one} implicit negation. An implicit negation is one that does not contain a negation operator. The word itself has negative semantics. Examples are ('avoid', 'refuse', 'deny', 'ignore'). It does \textbf{not} include affixal negations. The query should not contain any other negation. The query \textbf{must be well-defined and have a finite, verifiable answer} even outside the document. Avoid queries that could have an \textbf{infinite or unbounded} number of answers. The query must be specific, and sound like something someone would type into a search engine. \newline Step 2: Extract a short \textbf{retrieval-style passage} that answers the query. - In answering the query, the passage must contain exactly \textbf{the same} implicit negation as in the query. - If the passage \textbf{does not contain} the implicit negation, add it yourself. The passage should not contain any other negation.\\
			\hline
		\end{tabular}
        }
	\end{minipage}
		\caption{Summary of differences in prompt variants for different types of negation.}
		\label{tab:prompt-variants}
\end{figure*}

\header{Extra Verification for the generated instances} After generation, we filter the instances by prompting the LLM to check the relevance of the documents for the queries. We only keep the instances for which both pairs pass the relevance self-check. This verification step is needed as sometimes the generated queries are too general, making the retrieved document not highly relevant.

\begin{figure}[ht]
	\centering
	\includegraphics[width=0.75\linewidth]{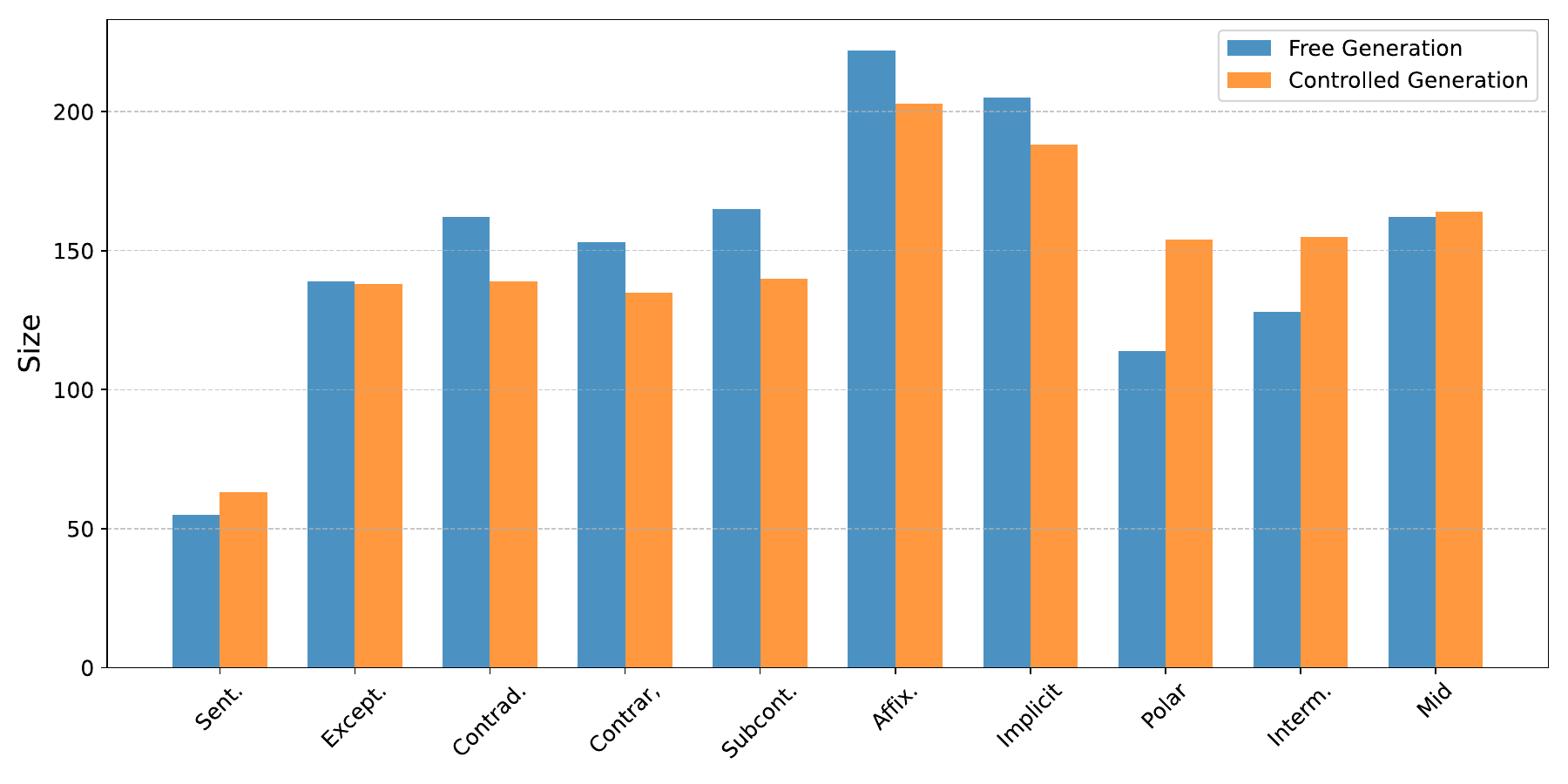}
	\caption{Distribution of negation types.}
	\label{fig:class_distr}
\end{figure}

\header{Label Distribution} Figure \ref{fig:class_distr} illustrates the distribution of negation types per synthetic dataset after the extra verification step. We notice that out of the generations, the sentential negations have been filtered the most. 

\header{Statistics of the generated datasets} Table \ref{tab:model_statistics} illustrates a summary of the two generated datasets, i.e., the free and controlled generation datasets. Length is calculated wrt. the number of words, while Data Size refers to the number of instances, where one instance is composed of pairs \({<}q_1, doc_1{>}\) and \({<}q_2, doc_2{>}\).

\begin{table}[ht]
  \centering
  \small           
  \setlength{\tabcolsep}{3pt}  
  \renewcommand{\arraystretch}{1.1} 
  \begin{tabular}{p{2.0cm} p{1.8cm} p{1.8cm}}
    \toprule
    \textbf{Statistics} & \textbf{Free Gen.} & \textbf{Contr. Gen.}  \\
    \midrule
    Data Size        & 1049/146/310   & 1031/143/305 \\ \hline
    Query1 length    & 10.25  & 10.20 \\
    Query2 length    & 10.82  & 10.60 \\ \hline
    Doc1 length      & 36.65  & 36.48 \\
    Doc2 length      & 33.35  & 33.26 \\
    \bottomrule
  \end{tabular}
	\caption{Statistics of the two generated datasets. Free Gen. stands for free generation dataset, while Controlled Gen. stands for controlled generation dataset.T The dataset size is split into partitions: train, validation, test.}
	\label{tab:model_statistics}
\end{table}

\subsection{What we do not cover}\label{appenfix:cool_negation_types}
This section contains negation phenomena and properties that, while interesting, we do not account for in this study.

\header{In scope non-negated events} These are examples of events that are not negated, despite being within the scope of a negation \citet{ConanDoyle2012negation}. Examples are shown below. We exclude these cases from our study.

\begin{itemize}[leftmargin=*]
    \item I should be glad to be able to say afterwards that I had solved it without [your help].
    \item I call it luck, but [it would] not [have come my way had I not been looking out for it].
    \item I call it luck, but it would not have come my way [had I] not [been looking out for it].
\end{itemize}

\header{Scope analysis} We also exclude analysis on the scope of the negation. In a sense, a query can be ``Restaurants that do not serve food'' and the returned document is ``Restaurants that do not wash laundry''. To maintain our study’s focus, we do not delve into scope considerations. Moreover, the scope of negation can often shift according to context. For example, negation can have outer-read and inner-reading, for example ``It is not likely that the Yankees will win.'':

\begin{itemize}[leftmargin=*]
    \item outer-reading: ~(Likely...) as in, it is not probable that it will happen that the Yankees will win. $\neg \exists $ 
    \item inner-reading: Likely ~ ... as in, it is likely the Yankees will not win. $\exists \neg x$
\end{itemize}

\header{Litotes} Double negation does not always reduce to x, i.e., not not x does not necessarily mean x \cite{horn2010multiple}. Such figure of speech is called a litotes, where an understatement is made by adding a negative. Example can be:

\begin{itemize}[leftmargin=*]
    \item \textit{I don't dislike cars.} ($\neg \forall \neg x = \exists \neg \neg x = \exists x$) can be seen as an understatement of \textit{I like cars.} ($\forall x$)
    \item \textit{Not bad!} is an understatement of \textit{Good!}
\end{itemize}


\header{Existential quantifiers with different scopes} Quantifiers such as ``every'' and ``some'' apply different scopes: Every man didn't win. Some man didn't win. $\forall x (\text{Man}(x) \rightarrow \neg W(x))$ and $\exists x (\text{Man}(x) \wedge \neg W(x))$.

\subsection{LM Logic classification}\label{appendix:lm_logic_classification}

When applying the typed lambda calculus formalization categorization, we check both pairs \((q_1, doc_2)\) and \((q_2, doc_1)\) for the presence of negation, as a result of not knowing necessarily where negation is present. For example, NevIR is constructed such that negation is always present in the first pair, while ExcluIR is constructed such that negation is always present in the second pair. Our classification mechanism is robust to such variations.

\begin{figure*}[!ht]
  \centering
  \begin{tcolorbox}[
    colback=white!95!black,
    colframe=black!75!white,
    width=\linewidth,
    title={System Prompt}
  ]
  \small{
    \begin{enumerate}[leftmargin=*]
      \item You are a Montagovian semanticist working in a typed $\lambda$-calculus framework.
      \item For each \textbf{input query}, follow the next four steps:
        \begin{enumerate}[label=\arabic*.]
          \item \textbf{LEXICON}: List every predicate and quantifier as a $\lambda$-term with an explicit Church type annotation.
          \item \textbf{SEMANTIC INVENTORY}: Output two comma-separated lists:
            \begin{itemize}[leftmargin=*]
              \item Predicates: \texttt{[]}
              \item Quantifiers: \texttt{[\(\exists, \forall\)]}
            \end{itemize}
          \item \textbf{NEGATION ANALYSIS}: For each predicate, indicate whether it matches one of the following categories:
            \begin{itemize}[leftmargin=*]
              \item Sentential (e.g.\ \textit{no}, \textit{not}, \textit{none}, \textit{never}, \textit{cannot})
              \item Exclusionary (e.g.\ \textit{besides}, \textit{except}, \textit{but})
              \item Affixal (e.g.\ bound morphemes \textit{im-}, \textit{in-}, \textit{un-}, \textit{-less}, etc.)
              \item Implicit (e.g.\ verbs such as \textit{deny}, \textit{refuse}, \textit{avoid}, \textit{fail})
            \end{itemize}
          \item \textbf{FINAL FORMULA}: Present the fully reduced $\lambda$-term for $S$, or an equivalent first- or higher-order logic formula, enclosed in a fenced code block.
        \end{enumerate}
      \item Respond in JSON format.
      \item \textbf{Example:}
        \begin{description}[style=nextline]
          \item[Query:] What organisms besides cyanobacteria perform anoxygenic photosynthesis?
            \item[LEXICON:] 
            organism: $\lambda x\!:\!e.\,\mathrm{Organism}(x)$, \\
            cyanobacteria: $\lambda x.\,\mathrm{Cyanobacteria}(x)$, \\
            perform\_anoxygenic\_photosynthesis: $\lambda x.\,\mathrm{PerformAnoxygenicPhotosynthesis}(x)$, \\
            besides: $\lambda P\,Q\,x.\,Q(x) \land \neg P(x)$
          \item[SEMANTIC INVENTORY:] Predicates: [Organism, Cyanobacteria, PerformAnoxygenicPhotosynthesis], Quantifiers: [\(\exists\)]
          \item[NEGATION ANALYSIS:] Sentential: [], Exclusionary: [besides], Affixal: [], Implicit: []
          \item[FINAL FORMULA:]
            \(\displaystyle
            \lambda x\colon e.\,\mathrm{Organism}(x)\;\land\;\mathrm{PerformAnoxygenicPhotosynthesis}(x)\;\land\;\neg\,\mathrm{Cyanobacteria}(x)
            \)
        \end{description}
    \end{enumerate}
    }
  \end{tcolorbox}
  \caption{Prompt for generating typed lambda calculus proofs.}
  \label{fig:typed_lambda_calculus}
\end{figure*}

\subsection{Annotators Template}\label{appendix:annotators}

The queries and documents have been shuffled within the instance, and the sample used for annotations has a perfectly balanced distribution of labels. Given an instance \((q_1, doc_1)\) and \((q_2, doc_2)\), we ask the following questions to the annotators:

\begin{enumerate}[label=\textbf{Q\arabic*:}, leftmargin=*]

  \item \textbf{Which document is more relevant for q1?}
    \begin{itemize}
      \item doc1
      \item doc2
      \item none
      \item both
    \end{itemize}

  \item \textbf{Which document is more relevant for q2?}
    \begin{itemize}
      \item doc1
      \item doc2
      \item none
      \item both
    \end{itemize}

  \item \textbf{Which instances contain negation? Multiple choices are possible.\\
            NOTE: If the individual instances do not contain negation, but the pair (q1, q2) contains antonyms, check both q1 and q2. Same goes for (doc1, doc2).}
      \begin{itemize}[label=\(\lozenge\), leftmargin=*]
        \item q1
        \item q2
        \item doc1
        \item doc2
      \end{itemize}

  \item \textbf{Rate the naturalness (fluency and readability) of the text.}
    \begin{enumerate}[label=\arabic*:, leftmargin=*]
      \item Text is forced 
      \item Noticeably awkward 
      \item Minor issues 
      \item Language flows well 
      \item Perfectly polished 
    \end{enumerate}

  \item \textbf{Rate the coherence (logical flow) of the text.}
    \begin{enumerate}[label=\arabic*:, leftmargin=*]
      \item No logical flow [e]
      \item Significant logical gaps 
      \item Basic logical structure 
      \item Generally logical and clear 
      \item Completely logical and clear 
    \end{enumerate}

  \item \textbf{Rate the consistency of information in the text.}
    \begin{enumerate}[label=\arabic*:, leftmargin=*]
      \item Contradictory 
      \item Unstable 
      \item Mixed 
      \item Aligned 
      \item Fully Aligned 
    \end{enumerate}
\end{enumerate}

\subsubsection{Statistical analysis on annotation results}\label{appendix:statistical_analysis}

\begin{table*}[!t]
  \centering
  \scriptsize
  \setlength{\tabcolsep}{3pt}
  \resizebox{\textwidth}{!}{%
    \begin{tabular}{l*{10}{r}}
      \toprule
      & \multicolumn{1}{c}{\textbf{T1}}
        & \multicolumn{1}{c}{\textbf{T2}}
        & \multicolumn{1}{c}{\textbf{T3}}
        & \multicolumn{1}{c}{\textbf{T4}}
        & \multicolumn{1}{c}{\textbf{T5}}
        & \multicolumn{1}{c}{\textbf{T6}}
        & \multicolumn{1}{c}{\textbf{T7}}
        & \multicolumn{1}{c}{\textbf{T8}}
        & \multicolumn{1}{c}{\textbf{T9}}
        & \multicolumn{1}{c}{\textbf{T10}} \\
      \midrule
    q1
    & \(0.79\pm0.21\)
    & \(0.64\pm0.21\)
    & \(0.79\pm0.07\)
    & \(0.71\pm0.14\)
    & \(0.86\pm0.00\)
    & \(0.79\pm0.07\)
    & \(0.79\pm0.07\)
    & \(0.79\pm0.07\)
    & \(0.79\pm0.07\)
    & \(0.64\pm0.21\) \\
    q2
    & \(0.79\pm0.07\)
    & \(0.21\pm0.07\)
    & \(0.93\pm0.07\)
    & \(0.71\pm0.00\)
    & \(0.79\pm0.07\)
    & \(0.79\pm0.07\)
    & \(0.71\pm0.00\)
    & \(0.79\pm0.07\)
    & \(0.79\pm0.07\)
    & \(0.57\pm0.14\)\\
    q3
    & \(0.91\pm0.04\)
    & \(1.00\pm0.00\)
    & \(0.90\pm0.04\)
    & \(0.96\pm0.03\)
    & \(0.94\pm0.01\)
    & \(0.87\pm0.03\)
    & \(0.90\pm0.08\)
    & \(0.81\pm0.00\)
    & \(0.77\pm0.14\)
    & \(0.69\pm0.07\) \\
    q4
    & \(3.86\pm0.00\)
    & \(3.71\pm0.37\)
    & \(4.29\pm0.57\)
    & \(3.79\pm0.21\)
    & \(4.21\pm0.21\)
    & \(4.29\pm0.14\)
    & \(4.07\pm0.18\)
    & \(4.36\pm0.07\)
    & \(4.21\pm0.07\)
    & \(4.29\pm0.29\) \\
    q5
    & \(3.86\pm0.14\)
    & \(4.21\pm0.24\)
    & \(4.07\pm0.36\)
    & \(3.57\pm0.14\)
    & \(4.14\pm0.00\)
    & \(4.29\pm0.14\)
    & \(4.14\pm0.14\)
    & \(4.29\pm0.00\)
    & \(4.21\pm0.21\)
    & \(4.07\pm0.21\) \\
    q6
    & \(3.86\pm0.29\)
    & \(4.21\pm0.26\)
    & \(4.50\pm0.50\)
    & \(4.57\pm0.14\)
    & \(4.29\pm0.00\)
    & \(3.71\pm0.57\)
    & \(3.79\pm0.36\)
    & \(4.50\pm0.36\)
    & \(3.79\pm0.79\)
    & \(3.93\pm0.36\)  \\
    \bottomrule
    \end{tabular}%
  }
    \caption{Performance of annotators with respect to the ground truth labels on the generated query-document pairs of both synthetically generated documents. Each score represents a mean with an std. error over the two datasets.}
    \label{tab:performance_annotators}
\end{table*}

\begin{table*}[!t]
  \centering
  \scriptsize
  \setlength{\tabcolsep}{3pt}
  \resizebox{\textwidth}{!}{%
    \begin{tabular}{l*{10}{r}}
      \toprule
      & \multicolumn{1}{c}{\textbf{T1}}
        & \multicolumn{1}{c}{\textbf{T2}}
        & \multicolumn{1}{c}{\textbf{T3}}
        & \multicolumn{1}{c}{\textbf{T4}}
        & \multicolumn{1}{c}{\textbf{T5}}
        & \multicolumn{1}{c}{\textbf{T6}}
        & \multicolumn{1}{c}{\textbf{T7}}
        & \multicolumn{1}{c}{\textbf{T8}}
        & \multicolumn{1}{c}{\textbf{T9}}
        & \multicolumn{1}{c}{\textbf{T10}} \\
      \midrule
      q1
        & \(0.60\pm0.02\)
        & \(0.26\pm0.17\) 
        & \(0.89\pm0.11\)
        & \(0.58\pm0.18\)
        & \(0.52\pm0.20\)
        & \(0.65\pm0.35\)
        & \(0.52\pm0.12\)
        & \(0.90\pm0.11\)
        & \(0.53\pm0.01\)
        & \(0.56\pm0.03\) \\
    q2
    & \(0.58\pm0.02\)
    & \(0.30\pm 0.02\) 
    & \(0.86\pm0.14\)
    & \(0.53\pm0.01\)
    & \(0.89\pm0.11\)
    & \(0.57\pm0.21\)
    & \(0.31\pm0.20\)
    & \(0.90\pm0.11\)
    & \(0.55\pm0.02\)
    & \(0.58\pm0.22\) \\
    q3
    & \(0.78\pm0.11\)
    & \(1.00\pm0.00\) 
    & \(0.93\pm0.01\)
    & \(1.00\pm0.00\)
    & \(0.92\pm0.08\)
    & \(0.74\pm0.16\)
    & \(0.67\pm0.08\)
    & \(0.85\pm0.05\)
    & \(0.87\pm0.13\)
    & \(0.87\pm0.02\) \\
    q4
    & \(0.80\pm0.01\)
    & \(0.30\pm0.20\) 
    & \(0.71\pm0.29\)
    & \(0.52\pm0.08\)
    & \(0.79\pm0.21\)
    & \(0.79\pm0.21\)
    & \(0.49\pm0.14\)
    & \(0.76\pm0.24\)
    & \(0.76\pm0.04\)
    & \(0.89\pm0.11\) \\
    q5
    & \(0.75\pm0.26\)
    & \(0.30\pm0.20\) 
    & \(0.68\pm0.32\)
    & \(0.63\pm0.37\)
    & \(0.89\pm0.11\)
    & \(0.76\pm0.02\)
    & \(0.69\pm0.10\)
    & \(0.64\pm0.09\)
    & \(0.71\pm0.29\)
    & \(0.37\pm0.01\) \\
    q6
    & \(0.55\pm0.02\)
    & \(0.36\pm0.30\) 
    & \(0.67\pm0.05\)
    & \(0.36\pm0.36\)
    & \(0.33\pm0.40\)
    & \(0.44\pm0.28\)
    & \(0.31\pm0.13\)
    & \(0.78\pm0.22\)
    & \(0.56\pm0.20\)
    & \(0.56\pm0.22\) \\
    \bottomrule
    \end{tabular}%
  }
    \caption{Inner Agreement of annotators on their answers about the generated query-document pairs of both synthetically generated documents. Each score represents a mean with an std. error over the two datasets.}
    \label{tab:inner_agreement}
\end{table*}

Table \ref{tab:performance_annotators} shows the performance of annotators with respect to the ground truth labels of the generated datasets, i.e., averaged over both the free and controlled generation datasets. The rows q1-q6 indicate the six questions presented to the annotators, and the columns T1-T10 present the results of their answers split across the ten types of negation present in the sample shown to the annotators. For a brief description of the questions: q1-q2 ask about the relevance of the two documents for each query, and are assessed through accuracy; q3 asks about the presence of negation in the generation (binary question; therefore, it does not ask about the specific \textit{type} of negation) and is assessed using the f1 score; q4-a6 are questions about the logic, naturalness, and consistency of information in the generated queries and documents, and are assessed by taking an average of the answers represented on an ordinal scale from 1-5.

Table \ref{tab:inner_agreement} shows the inner agreement of the annotators when answering the questions wrt. the two generated datasets, i.e., averaged over both the free and controlled generation datasets. The rows q1-q6 indicate the six questions presented to the annotators, and the columns T1-T10 present the results of their answers split across the ten types of negation present in the sample shown to the annotators. For a brief description of the questions: q1-q2 ask about the relevance of the two documents for each query, and the agreement is measured using Cohen's Kappa; q3 asks about the presence of negation in the generation (binary question; therefore, it does not ask about the specific \textit{type} of negation) and is assessed using recall of agreement; q4-a6 are questions about the logic, naturalness, and consistency of information in the generated queries and documents, and are assessed using a weighted Cohen's Kappa, given the answers represent an ordinal scale from 1-5. The scores are averaged across the two datasets.

\subsection{Results}\label{ap:results}

\begin{figure*}[ht] 
	\centering
	\includegraphics[width=0.8\linewidth]{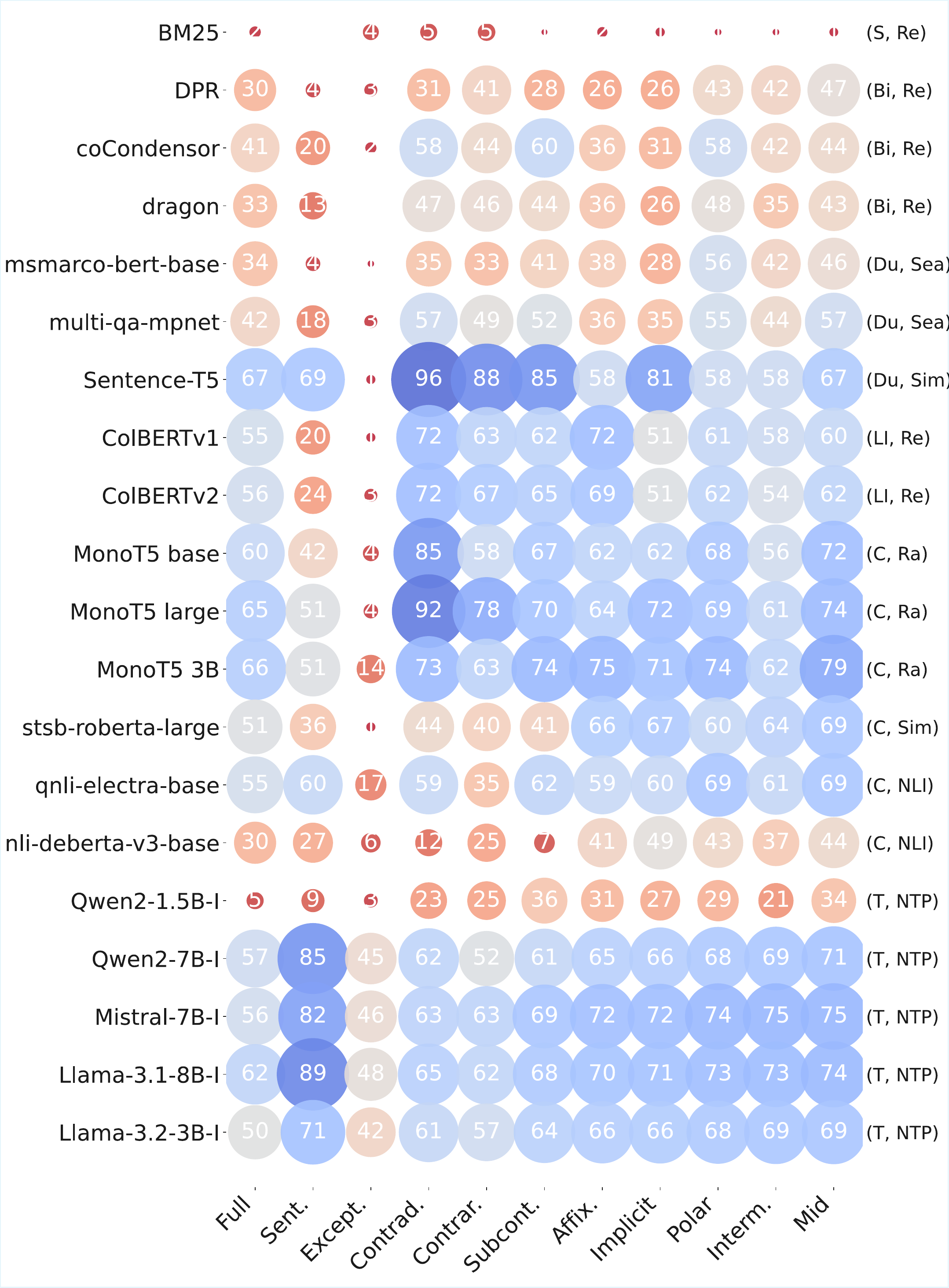}
	\caption{Close-up of results on the Free Generation.}
	\label{fig:free_gen_pairwise_acc_close_up}
\end{figure*}

\begin{figure*}[ht]
	\centering
	\includegraphics[width=1\linewidth]{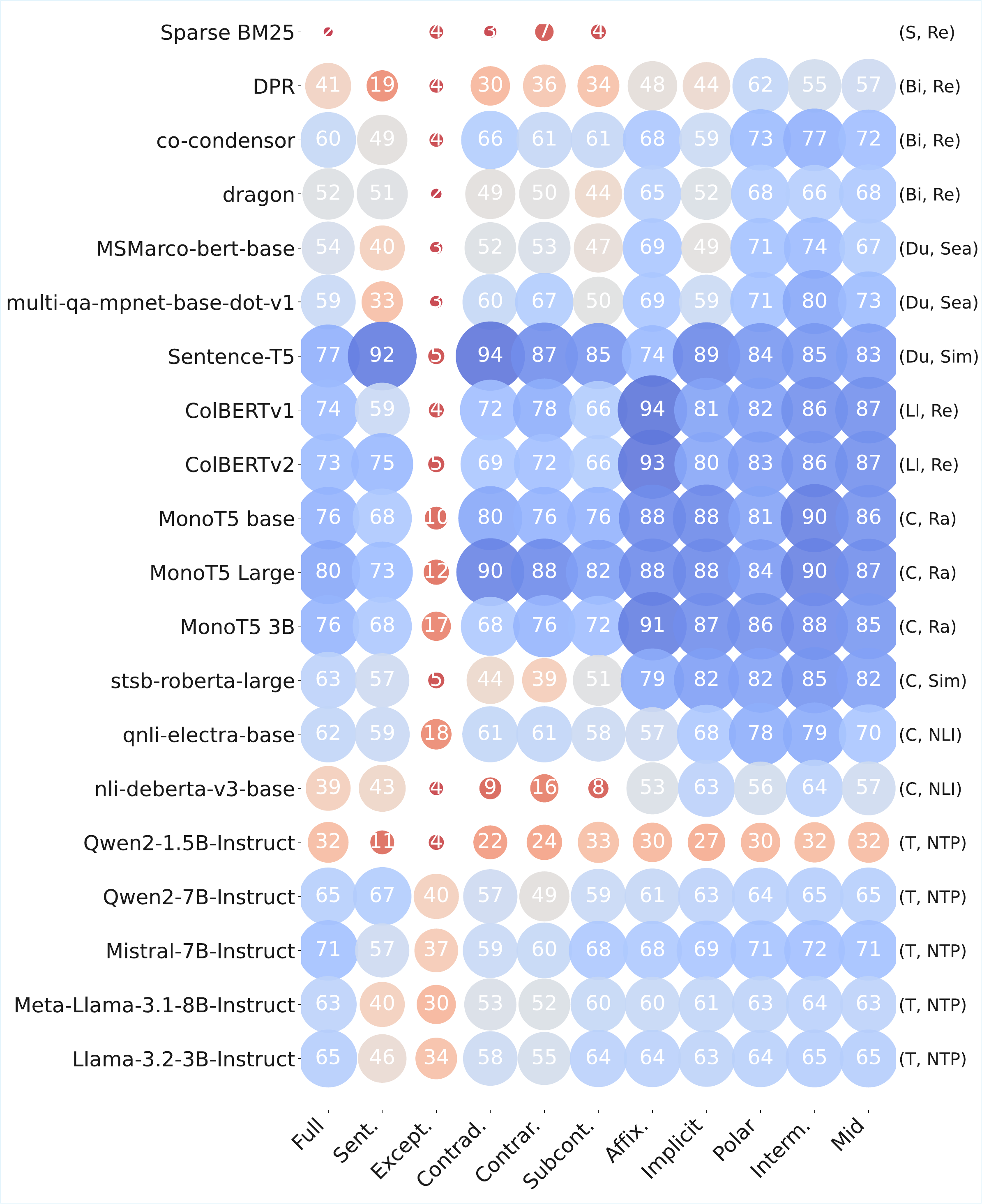}
	\caption{Pairwise Accuracy on the Controlled Generations dataset. Each column represents a negation type following our taxonomy, including the Full dataset in the first column. Each model is represented by one row.}
	\label{fig:controlled_gen_pairwise_acc}
\end{figure*}

\begin{figure*}[ht]
	\centering
	\includegraphics[width=0.6\linewidth]{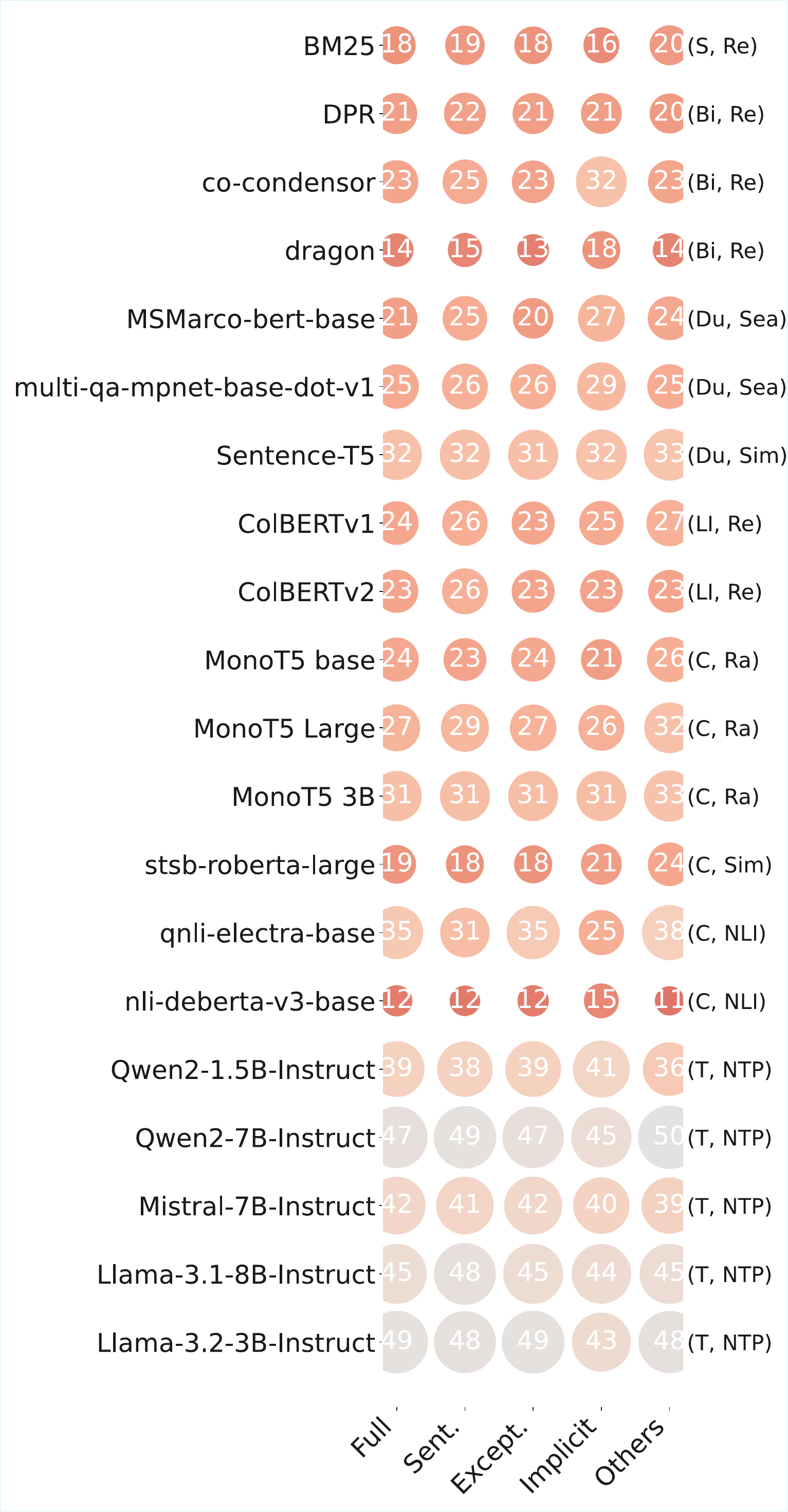}
	\caption{Pairwise Accuracy on ExcluIR. The dataset is split with out classification Mechanism.}
	\label{fig:excluirpa}
\end{figure*}

In Figures \ref{fig:free_gen_pairwise_acc_close_up}, \ref{fig:controlled_gen_pairwise_acc} and \ref{fig:excluirpa} we illustrate a close-up of the free generation synthetic experiments, the controlled generation experiments, and evaluation on ExcluIR as a result of our categorization mechanism.

\vspace{5mm}

\clearpage
\subsubsection{Finetuning curves}\label{appendix:finetuning}

\begin{figure}[ht]
	\centering
	\includegraphics[width=0.6\linewidth]{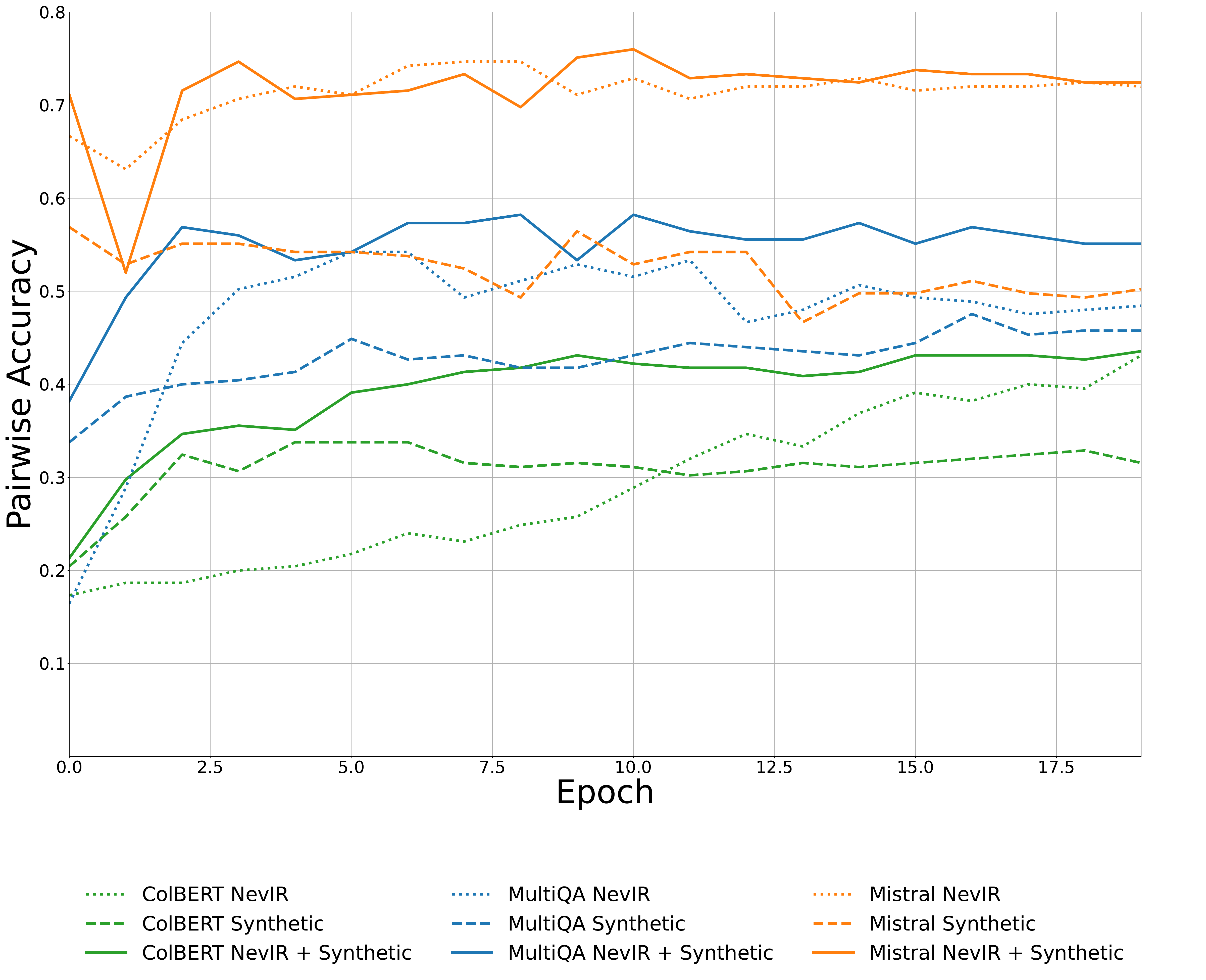}
	\caption{Fine-tuning results for ColBERT and MultiQA on 3 datasets: NevIR train, free generation train, and Mixed. Evaluated against NevIR dev.}
	\label{fig:finetuning_nevir}
\end{figure}

\begin{figure}[ht]
	\centering
	\includegraphics[width=0.6\linewidth]{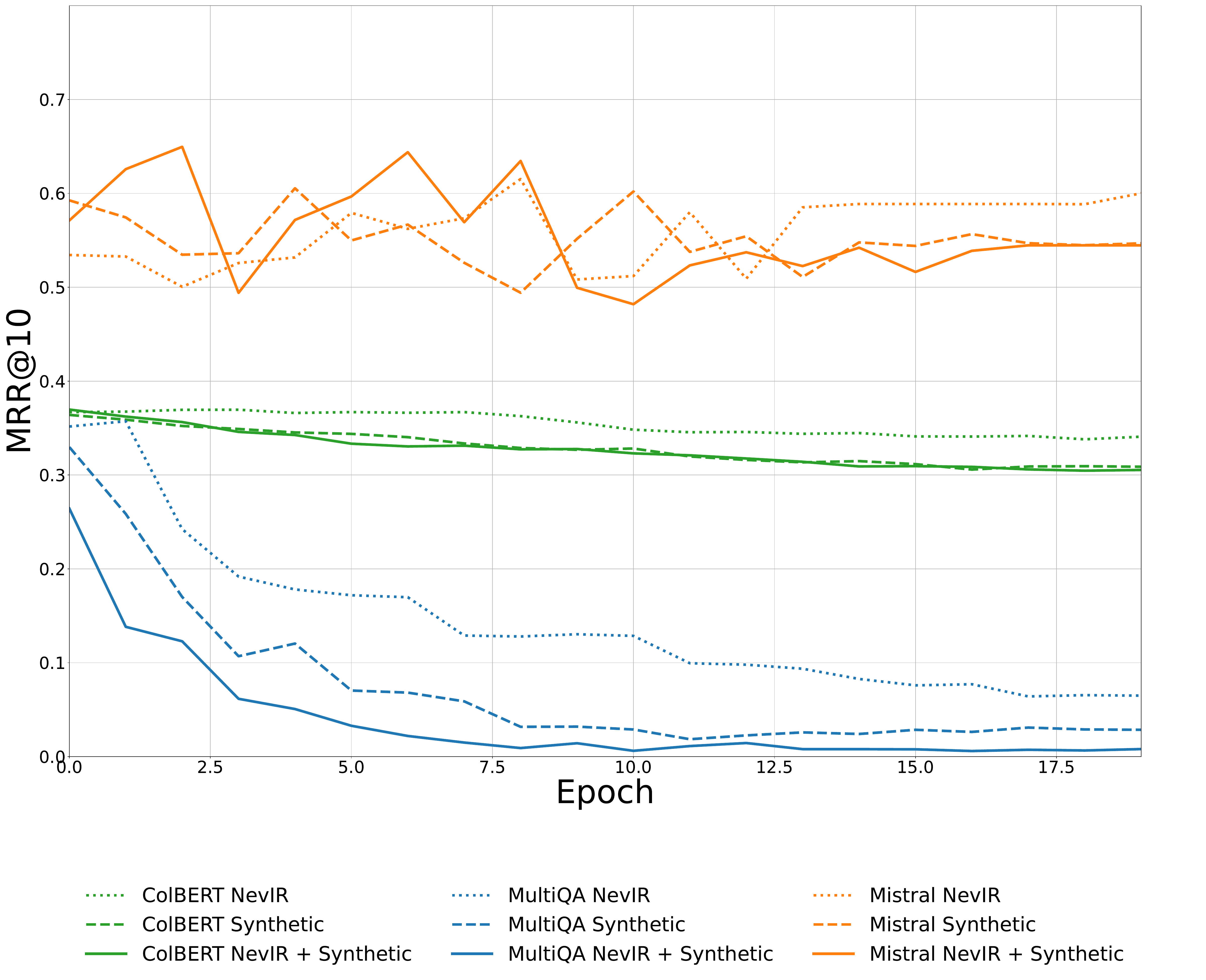}
	\caption{Fine-tuning results for ColBERT and MultiQA on 3 datasets: NevIR train, free generation train, and Mixed. Evaluated against MSMarco dev.}
	\label{fig:colbert_finetuning_eval_msmarco}
\end{figure}

Figures \ref{fig:finetuning_nevir} and \ref{fig:colbert_finetuning_eval_msmarco} illustrate the fine-tuning curves for ColBERT, MultiQA and Mistral when fine-tuned on synthetic, NevIR, and a mix of the two datasets. The evaluation is done on NevIR with pairwise accuracy, and on MSMarco with MRR@10.

\end{document}